\documentclass[letterpaper]{article} 
\usepackage[preprint]{aaai2027}
\usepackage[hyphens]{url}  
\usepackage{graphicx} 
\usepackage{natbib}  
\usepackage{caption} 

\usepackage{algorithm}
\usepackage{algorithmic}
\usepackage{booktabs}
\usepackage{multirow}
\usepackage{amsmath}
\usepackage{amssymb}
\usepackage{amsthm}
\newtheorem{theorem}{Theorem}
\newtheorem{remark}{Remark}
\usepackage{subcaption}

\usepackage{array}
\usepackage[table]{xcolor} 

\usepackage{newfloat}
\usepackage{listings}
\DeclareCaptionStyle{ruled}{labelfont=normalfont,labelsep=colon,strut=off} 
\floatstyle{ruled}
\newfloat{listing}{tb}{lst}{}
\floatname{listing}{Listing}

\usepackage{booktabs}

\title{
CRMSurv: Learning Compact and Risk-discriminative Representations for Multimodal Cancer Survival Analysis

}

\author{
    Huayi Wang\textsuperscript{\rm 1,2,6,7},
    Yutao Sun\textsuperscript{\rm 1,3,6,7},
    Wenjie Deng\textsuperscript{\rm 1,2,6,7},
    Yue Wang\textsuperscript{\rm 1,2,6,7}, \\
    Haochao Ying\textsuperscript{\rm 1,4}\corresponding,
    Jian Wu\textsuperscript{\rm 1,5,6,7}
}

\affiliations{
    \textsuperscript{\rm 1}State Key Laboratory of Transvascular Implantation Devices, Second Affiliated Hospital of Zhejiang University School of Medicine\\
    \textsuperscript{\rm 2}College of Artificial Intelligence, Zhejiang University\\
    \textsuperscript{\rm 3}Polytechnic Institute, Zhejiang University\\
    \textsuperscript{\rm 4}School of Public Health, Zhejiang University\\
    \textsuperscript{\rm 5}College of Computer Science and Technology, Zhejiang University \\
    \textsuperscript{\rm 6}Transvascular Implantation Devices Research Institute \\
    \textsuperscript{\rm 7}Zhejiang Key Laboratory of Medical Imaging Artificial Intelligence \\
    \{huayiwang, yutaosun, 12521290, ywang2022, haochaoying, wujian2000\}@zju.edu.cn
}

\begin{document}

\maketitle

\begin{abstract}
Multimodal survival analysis has substantially advanced prognosis prediction by integrating heterogeneous data such as pathology, genomics, and textual reports. 
While existing studies primarily focus on cross-modal fusion, effective analysis also depends on the quality of the modality-specific representations learned prior to fusion.
Two challenges remain. 
First, massive whole slide image (WSI) patches with high-frequency redundancies impose a severe computational burden, making compression essential. 
However, existing methods often achieve representation compactness at the cost of discarding locally informative evidence.
Second, small-scale cohort datasets are insufficient to support the training of high-dimensional multimodal representations, leaving models prone to overfitting prognosis-irrelevant noise.
To address these issues, we propose CRMSurv, a novel framework that learns \underline{C}ompact and \underline{R}isk-discriminative representations for \underline{M}ultimodal cancer \underline{Surv}ival analysis. 
Specifically, we develop Reconstructability-aware Patch (RP) Compression. By combining compact prototypes and high-reconstruction-error patches, it optimally balance primary morphology and complementary local details with fewer tokens.
Furthermore, we introduce patient-level Risk-discriminative Learning (RL) with a dynamic risk prototype bank to explicitly supervise representation learning, which mitigates overfitting by anchoring representations to stable prognostic patterns.
Finally, we adopt multi-branch Q-Formers and cross-attention to align these refined representations for multimodal fusion.
On six cancer datasets, CRMSurv achieves a state-of-the-art average C-index of 0.728 while retaining just 25\% of WSI patches.
\end{abstract}

\section{Introduction}

Cancer survival analysis
aims to assess patients’ mortality risk and prognosis over time, supporting risk stratification and treatment decision-making~\cite{Giordano_2021,sox2024medical}. 
Multimodal learning further improves prognosis prediction by integrating heterogeneous and complementary data, such as pathology images, genomic profiles, and textual reports~\cite{Boehm_2022,Sabah_2021}. 
To effectively exploit multimodal complementary information, most existing studies mainly focus on designing cross-modal fusion and interaction mechanisms, including feature concatenation~\cite{hao2019page}, attention-based interaction~\cite{MCAT}, and mixture-of-experts-based interaction~\cite{xiong2024mome}. 
However, effective multimodal learning depends not only on these interaction strategies, but also on the quality of the pre-interaction representations. 
Ideally, a set of compact, information-preserving, and prognostic representations can better align complementary evidence, thereby minimizing interference from irrelevant noise.
Nevertheless, constructing such representations remains challenging in two aspects.
\begin{figure}
    \centering
    \includegraphics[width=\linewidth]{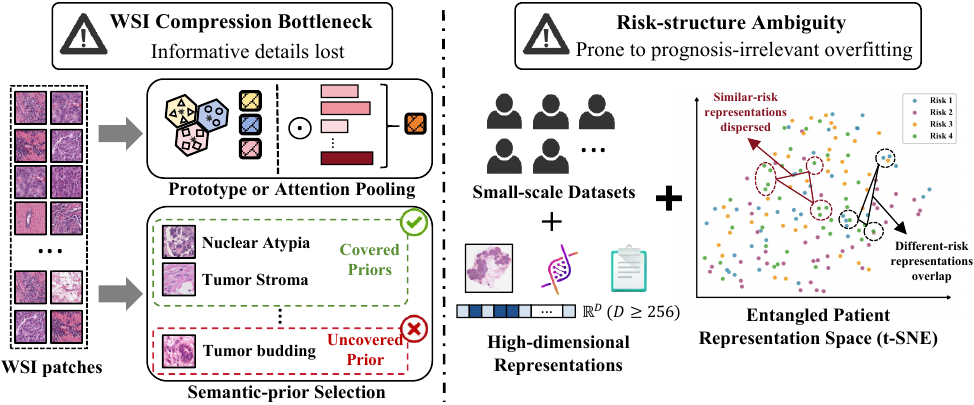}
    \caption{Challenges in Pre-interaction Representations. Left: Existing WSI compression methods struggle to balance compactness with complementary prognostic details. 
    Right: Limited samples, high-dimensional inputs, and ambiguous risk structures jointly increase the risk of overfitting.
    }
    \label{fig:intro}
    \vspace{-0.2cm}
\end{figure}

\textbf{\textit{Compressing massive WSI patches without sacrificing informative local details presents a fundamental dilemma.}} As shown in Figure~\ref{fig:intro} (left), under the classic multiple instance learning (MIL) paradigm~\cite{AttnMIL}, each WSI is represented by massive patches ($>10^4$), resulting in extremely large inputs. Such representations incur high computational costs and overwhelm sparse prognostic signals~\cite{buzzard2024paths}, disrupting subsequent cross-modal interaction with irrelevant visual noise. 
Existing methods reduce the number of patches through attention pooling~\cite{Boehm_2022}, prototype-based compression~\cite{zhou2024cohort,Raza2025PS3AM}, or semantic-prior-based patch selection~\cite{PAMoE,Huang2026ActLA}.
However, attention pooling and prototype-based methods overcompress WSI representations into a few global summaries, in which localized risk-informative patterns are inevitably smoothed out by dominant tissue morphology.
Semantic-prior-based selection, on the other hand, is constrained by predefined textual concepts and may overlook informative regions beyond their semantic coverage.

\textbf{\textit{Learning robust prognostic representations from small-scale cohort datasets is challenging.}} Due to privacy and data acquisition costs, public cancer survival datasets remain small-scale~\cite{lipkova2022artificial}. 
Meanwhile, multimodal representations are inherently high-dimensional, making models trained on limited cohorts susceptible to overfitting~\cite{berisha2021digital}. 
Furthermore, by visualizing modality-specific representations (Figure~\ref{fig:intro}, right), we observe a poorly organized space where patients with similar risks remain dispersed while those with different risks overlap. This phenomenon indicates that models fail to capture prognosis-relevant signals, as robust representations should naturally cluster patients according to their true risk labels. Together, the combination of limited cohort sizes, high-dimensional representations, and ambiguous risk structures makes models prone to fitting prognosis-irrelevant signals instead of learning stable prognostic patterns.



To address the above issues, we propose CRMSurv, a novel framework for learning \underline{C}ompact and \underline{R}isk-discriminative representations for \underline{M}ultimodal cancer \underline{Surv}ival analysis. 
First, we develop Reconstructability-aware Patch (RP) Compression to improve the compactness and information preservation of WSI representations.
From an information-theoretic perspective, we prove that the conditional uncertainty of the raw WSI given its compressed tokens is upper-bounded by a reconstruction error function. 
Guided by this, RP Compression uses compact prototypes as basis tokens to reconstruct all WSI patches via least squares, retaining high-error patches to complement information unexplained by prototypes. Combining these retained patches with prototypes effectively reduces the uncertainty bound, preserving dominant morphology and fine-grained local evidence at a lower cost.
Second, we introduce patient-level Risk-discriminative Learning (RL) with a dynamic risk prototype bank to supervise the model in identifying prognosis-relevant modality representations. This bank maintains risk-group prototypes to provide explicit risk-aware supervision. By promoting intra-risk consistency and inter-risk separation, RL organizes patient-level representations around stable prognostic patterns, mitigates overfitting to irrelevant noise, and improves generalization.
After obtaining optimized modality representations, we employ multi-branch Q-Formers~\cite{Li2023BLIP2BL} and cross-attention to adapt heterogeneous modality features and align their complementary information for multimodal fusion.
The main contributions are summarized as follows:

\begin{enumerate}
    \item We propose CRMSurv, a novel multimodal survival framework that targets compact, information-preserving, and prognosis-oriented representation learning.

    \item We introduce RP Compression to preserve informative pathology patterns with fewer patches, and patient-level RL with dynamic risk prototypes to supervise prognosis-relevant modality representation learning.

    \item 
    Extensive experiments on six cancer datasets show that CRMSurv achieves a state-of-the-art average C-index of 0.728, even when retaining only 25\% of WSI patches.
\end{enumerate}

\section{Related Works}
\subsection{Multimodal Survival Analysis}
Existing methods mainly focus on cross-modal representation fusion and interaction modeling. For example, concatenation-based methods~\cite{hao2019page} directly concatenate features from different modalities for joint prediction; attention-based methods~\cite{MCAT,CMTAs,Raza2025PS3AM} model the associations between pathology and genomic features; and MoE-based method~\cite{xiong2024mome} further capture optimal multimodal fusion ways through dynamic expert selection. In addition, recent studies~\cite{wang2025decouple,zhou2024cohort} have explored decoupling-fusion paradigms, which improve multimodal fusion by decoupling and recombining modality-shared and modality-specific information between modalities. 
However, most of these methods primarily focus on pathology and genomic modalities, underexploiting the rich semantic information available in the associated text reports.
Moreover, existing multimodal fusion methods often aggregate all patches with an attention-based CLS token or represent WSIs using only a few prototypes. Such over-compression inevitably dilutes informative local details.

\subsection{WSI Patch Selection and Compression}
In computational pathology and MIL, many studies have explored how to select or compress WSI patches before slide-level prediction. For example, attention-pooling-based methods~\cite{AttnMIL,CLAM} aggregate patches according to attention weights, but they typically require additional parameters to learn token importance. 
Prototype-based methods~\cite{PIBD,jaume2024modeling,song2024mmp} replace patch tokens with prototype representations, but may lose fine-grained information during compression. Another line of methods~\cite{Huang2026ActLA,PAMoE} selects patches using predefined histological concepts priors; however, their effectiveness is limited by the coverage of these priors, which may cause potentially prognostic regions to be overlooked. 
Finally, Guo et al.~\cite{guo2025focus} propose a sliding-window-based patch selection algorithm. However, it requires similarity computation within each local window. Since WSIs contain massive patches, this process needs to be repeated many times, leading to heavy computational burden and longer processing time.
Different from these methods, we propose RP Compression. It requires no additional parameter learning or manually defined semantic priors. Instead of replacing WSI patches with only prototypes, RP Compression uses prototype-based reconstructability to retain hard-to-reconstruct informative patches, and combines them with prototype summaries to reduce patch numbers while preserving local details.
\section{Methodology}
\begin{figure*}
    \centering
    \includegraphics[width=\linewidth]{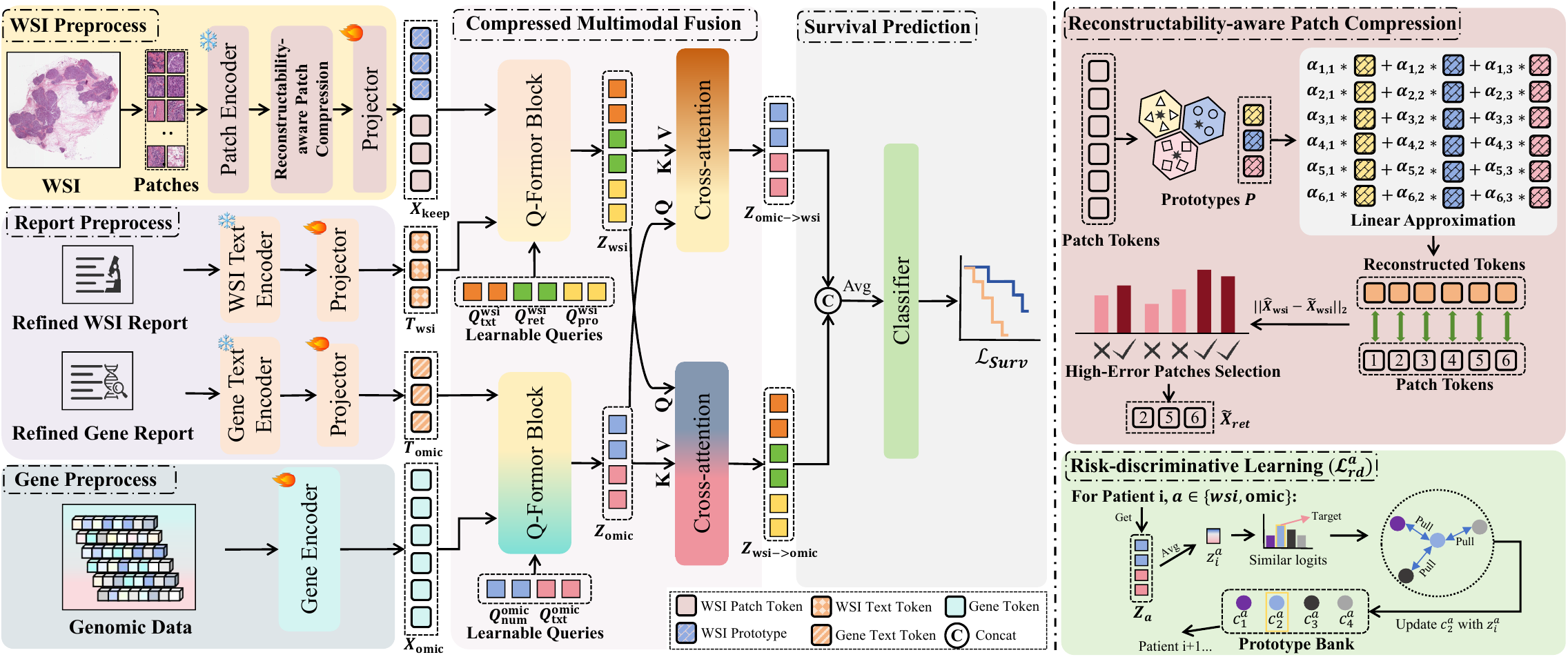}
    \caption{Overview of CRMSurv. Left: Data Preprocessing, Compressed  Multimodal Fusion, and Survival Prediction. Right: Reconstructability-aware Patch (RP) Compression and Risk-discriminative Learning.}
    \label{fig:framework}
\end{figure*}
Our CRMSurv Framework is shown in Figure~\ref{fig:framework}. It consists of a Reconstructability-aware WSI Patch Compression module, a Compressed Multimodal Fusion module, and an additional Risk-discriminative Learning loss. CRMSurv integrates WSI images, genomic data, and their corresponding textual reports for cancer survival analysis.

\subsection{Data Preparation}
For each patient, we denote the WSI patch features as 
$\tilde{X}_{\mathrm{wsi}}=\{x^w_1,x^w_2,\ldots,x^w_n\}\in \mathbb{R}^{n\times d_w}$,
and the refined WSI report features as
$\tilde{T}_{\mathrm{wsi}}=\{t^w_1,t^w_2,\ldots,t^w_m\}\in \mathbb{R}^{m\times d_t}$.
For the genomic modality, the numerical features and report features are denoted as
$X_{\mathrm{omic}}=\{x^o_1,x^o_2,\ldots,x^o_6\}\in \mathbb{R}^{6\times d}$
and
$\tilde{T}_{\mathrm{omic}}=\{t^o_1,t^o_2,\ldots,t^o_l\}\in \mathbb{R}^{l\times d_g}$,
respectively.
Here, $n$ is the number of WSI patches, $m$ and $l$ are the numbers of WSI and genomic report tokens, and $d_w$, $d_t$, $d_g$, and $d$ denote the corresponding feature dimensions.
Before multimodal fusion, all representations are aligned to a unified hidden dimension $d$.
After alignment and WSI compression, we obtain four inputs:
$X_{\mathrm{keep}}=\{X_{\mathrm{ret}}\cup P\}\in\mathbb{R}^{(k+r)\times d}$,
$X_{\mathrm{omic}}\in\mathbb{R}^{6\times d}$,
$T_{\mathrm{wsi}}\in\mathbb{R}^{m\times d}$, and
$T_{\mathrm{omic}}\in\mathbb{R}^{l\times d}$,
where $k$ and $r$ denote the numbers of retained patches and compact prototypes, respectively.
The detailed data construction and refinement processes are provided in Appendix.

\subsection{Reconstructability-aware Patch Compression}
A large patch numbers in WSIs introduce additional computational burden and weaken model’s attention to key prognostic regions. Existing studies have attempted to compress WSI patches using attention pooling, prototypes, or semantic-prior-based strategies. However, these methods either introduce additional learnable parameters or risk discarding informative details during compression. To address this, we seek to reduce WSI patches while preserving essential information, and propose a Reconstructability-aware Patch (RP) Compression to measure the irreplaceability of each patch from an information-theoretic perspective.

Given the WSI patch feature set $\tilde{X}_{\mathrm{wsi}}$, we aim to obtain a compact subset $\tilde{X}_{\mathrm{keep}}\subseteq \tilde{X}_{\mathrm{wsi}}$ that preserves as much information as possible from the raw WSI representation. This objective can be formulated as maximizing their mutual information, i.e., $\tilde{X}_{\mathrm{keep}}^*=\arg\max I(\tilde{X}_{\mathrm{wsi}};\tilde{X}_{\mathrm{keep}})$. According to the definition of mutual information~\cite{mackay2003information,ma2026apet}, $I(\tilde{X}_{\mathrm{wsi}};\tilde{X}_{\mathrm{keep}})=H(\tilde{X}_{\mathrm{wsi}})-H(\tilde{X}_{\mathrm{wsi}}\mid \tilde{X}_{\mathrm{keep}})$. Since $H(\tilde{X}_{\mathrm{wsi}})$ is fixed, maximizing $I(\tilde{X}_{\mathrm{wsi}};\tilde{X}_{\mathrm{keep}})$ is equivalent to minimizing $H(\tilde{X}_{\mathrm{wsi}}\mid \tilde{X}_{\mathrm{keep}})$, i.e., the remaining uncertainty of the raw WSI representation after compression.
However, directly estimating $H(\tilde{X}_{\mathrm{wsi}}\mid \tilde{X}_{\mathrm{keep}})$ is intractable in the high-dimensional WSI feature space. 
We therefore propose and prove that the reconstruction error can  provide a computable bound-related surrogate for the conditional uncertainty $H(\tilde{X}_{\mathrm{wsi}}\mid \tilde{X}_{\mathrm{keep}})$, i.e., a lower global reconstruction error reduces the upper bound of $H(\tilde{X}_{\mathrm{wsi}}\mid \tilde{X}_{\mathrm{keep}})$ (Proof 1 in Appendix).
Following this principle, we start from an initialized compact prototype set and add high-error patches that are poorly reconstructed by these prototypes, thereby reducing the global reconstruction error and further lowering the conditional uncertainty.

Specifically, we adopt KMeans to construct $r$ compact prototypes from the raw WSI patches:
$\tilde{P}=\{p_1,p_2,\ldots,p_r\}\in \mathbb{R}^{r\times d_w}$,
where $r\ll n$.
These prototypes summarize frequent morphological patterns in WSIs, such as tumor regions, stromal components, inflammatory infiltration, and other common tissue structures.
Then, for each patch token $x_i^w$, we use $\tilde{P}$ as basis tokens and perform least-squares linear reconstruction, which provides a simple non-iterative solution without additional trainable parameters:
\begin{equation}
    \hat{x}_i^w\approx\sum_{j=1}^{r}\alpha_{ij}p_j,
    \label{eq:linear}
\end{equation}
where $\alpha_{ij}$ denotes the contribution of the $j$-th prototype to reconstructing $x_i^w$. 
The reconstruction error is computed as:
\begin{equation}
    e_i=\|x_i^w-\hat{x}_i^w\|_2.
    \label{eq:mse}
\end{equation}
This error can be regarded as a measure of patch-token irreplaceability. 
A smaller $e_i$ indicates that the patch can be well represented by the prototypes. Conversely, a larger $e_i$ indicates that the patch is difficult to reconstruct and may contain additional pathology information beyond the dominant prototype patterns. Finally, we sort all patch tokens by their reconstruction errors and retain the Top-$k$ high-error patches as complementary details to the prototypes, formulated as $\tilde{X}_{\mathrm{keep}}=\{\tilde{X}_{\mathrm{ret}}\cup\tilde{P}\}\in \mathbb{R}^{(k+r)\times d_w}$.
In this way, $\tilde{P}$ preserves the dominant reconstructable information, while $\tilde{X}_{\mathrm{ret}}$ supplements the residual hard-to-reconstruct information, jointly reducing the remaining uncertainty of the raw WSI representation.


\subsection{Compressed Multimodal Fusion}
Given the inherent distributional differences among pathology, genomic, and textual modalities, direct fusion may cause feature misalignment and semantic noise.
Meanwhile, genomic features and compressed WSI features may still contain signals weakly related to survival prediction, necessitating a task-relevant information bottleneck~\cite{nagrani2021bottleneck} for further filtering.
Therefore, we design two multi-branch Q-Former~\cite{Li2023BLIP2BL} modules to first bridge the heterogeneous modality-specific features and their corresponding textual semantics.
In each branch, $h$ learnable query tokens serve as bottlenecks to extract compact prognosis-relevant representations from the corresponding input source.
Then, bidirectional cross-attention is employed to align pathology and genomics while capturing complementary prognostic information between them.

For the pathology modality, we obtain $X_{\mathrm{ret}}\in\mathbb{R}^{k\times d}$, $P\in\mathbb{R}^{r\times d}$, and $T_{\mathrm{wsi}}\in\mathbb{R}^{m\times d}$ after projection. Since these three sources correspond to different levels of pathology information, we assign three groups of learnable queries to them, namely
$Q^{\mathrm{wsi}}_{\mathrm{ret}}$, $Q^{\mathrm{wsi}}_{\mathrm{pro}}$, and $Q^{\mathrm{wsi}}_{\mathrm{txt}}$.
Here, $Q^{\mathrm{wsi}}_{\mathrm{ret}}$ focuses on high-value retained local patches, $Q^{\mathrm{wsi}}_{\mathrm{pro}}$ summarizes the compact morphological patterns represented by prototypes, and $Q^{\mathrm{wsi}}_{\mathrm{txt}}$ extracts high-level pathology semantics from the textual features. We denote the query-based feature extraction operation in Q-Former as $\mathcal{E}(\cdot,\cdot)$. The pathology-specific query representations are obtained as:
\begin{equation}
\begin{aligned}
\hat{Q}^{\mathrm{wsi}}_{\mathrm{ret}}
&=\mathcal{E}(Q^{\mathrm{wsi}}_{\mathrm{ret}}, X_{\mathrm{ret}}), \quad
\hat{Q}^{\mathrm{wsi}}_{\mathrm{pro}}
=\mathcal{E}(Q^{\mathrm{wsi}}_{\mathrm{pro}}, P), \\
\hat{Q}^{\mathrm{wsi}}_{\mathrm{txt}}
&=\mathcal{E}(Q^{\mathrm{wsi}}_{\mathrm{txt}}, T_{\mathrm{wsi}}).
\end{aligned}
\end{equation}
Then, the three query groups are concatenated and integrated by a self-attention block:
\begin{equation}
    Z_{\mathrm{wsi}}
    =
    \operatorname{FFN}
    \left(
    \operatorname{SelfAttn}
    \left(
    [\hat{Q}^{\mathrm{wsi}}_{\mathrm{ret}};
    \hat{Q}^{\mathrm{wsi}}_{\mathrm{pro}};
    \hat{Q}^{\mathrm{wsi}}_{\mathrm{txt}}]
    \right)
    \right).
\end{equation}
In this way, the representation $Z_{\mathrm{wsi}}$ jointly encodes retained patches, prototype-based summaries of unretained patches, and report-level textual semantics in a unified query space.

For the genomic modality, we obtain genomic numerical features $X_{\mathrm{omic}}\in\mathbb{R}^{6\times d}$ and genomic textual features $T_{\mathrm{omic}}\in\mathbb{R}^{l\times d}$ after projection. 
Accordingly, two query groups $Q^{\mathrm{omic}}_{\mathrm{num}}$ and $Q^{\mathrm{omic}}_{\mathrm{txt}}$ are used to extract molecular numerical patterns and genomic textual semantics, respectively.
The final genomic-specific query representations are obtained as:
\begin{equation}
\hat{Q}^{\mathrm{omic}}_{\mathrm{num}}
=\mathcal{E}(Q^{\mathrm{omic}}_{\mathrm{num}}, X_{\mathrm{omic}}), 
\hat{Q}^{\mathrm{omic}}_{\mathrm{txt}}
=\mathcal{E}(Q^{\mathrm{omic}}_{\mathrm{txt}}, T_{\mathrm{omic}}).
\end{equation}
Then, the two query groups are concatenated and integrated by the self-attention block:
\begin{equation}
    Z_{\mathrm{omic}}
    =
    \operatorname{FFN}
    \left(
    \operatorname{SelfAttn}
    \left(
    [\hat{Q}^{\mathrm{omic}}_{\mathrm{num}};
    \hat{Q}^{\mathrm{omic}}_{\mathrm{txt}}]
    \right)
    \right).
\end{equation}
In this way, the representation $Z_{\mathrm{omic}}$ jointly encodes pathway-level molecular patterns and report-level mechanism semantics in a unified query space. The detailed Q-Former blocks are provided in Appendix.

Later, we employ bidirectional cross-attention to exchange complementary information between pathology and genomics. This allows the pathology representation to incorporate molecular cues from genomic representation, while enabling the genomic representation to capture morphology-related context from pathology representation:
\begin{equation}
\begin{aligned}
Z_{\mathrm{wsi}\rightarrow \mathrm{omic}}
&=\operatorname{CrossAttn}(Z_{\mathrm{wsi}}, Z_{\mathrm{omic}}, Z_{\mathrm{omic}}),\\
Z_{\mathrm{omic}\rightarrow \mathrm{wsi}}
&=\operatorname{CrossAttn}(Z_{\mathrm{omic}}, Z_{\mathrm{wsi}}, Z_{\mathrm{wsi}}).
\end{aligned}
\end{equation}
The two enhanced representations are concatenated and mean-pooled to obtain the fused patient-level representation:
\begin{equation}
    Z_{\mathrm{mean}}
    =
    \operatorname{MeanPool}
    \left(
    [Z_{\mathrm{wsi}\rightarrow \mathrm{omic}};
    Z_{\mathrm{omic}\rightarrow \mathrm{wsi}}]
    \right)
    \in \mathbb{R}^{d}.
\end{equation}
Finally, $Z_{\mathrm{mean}}$ is fed into a linear classifier for discrete-time survival prediction.

\subsection{Risk-discriminative Learning}

\begin{algorithm}[t]
\caption{Risk-discriminative Learning for One Modality}
\label{alg:risk_discriminated_learning}

\begin{algorithmic}[1]
\REQUIRE Modality representation $Z_i^a$, risk label $y_i$, censoring indicator $c_i$, prototype bank $C^a=\{c_g^a\}_{g=1}^{G}$
\ENSURE Risk-discriminative loss $\mathcal{L}_{\mathrm{rd}}^{a}$ and updated prototype bank $C^a$

\STATE Initialize $C^a$ as zero vectors at the beginning of training: $c_g^a=\mathbf{0},\ g=1,\ldots,G$.

\STATE Obtain patient-level representation by average pooling:
$z_i^a=\operatorname{Norm}(\frac{1}{h}\sum_{j=1}^{h} Z_{i,j}^{a})$.

\STATE Compute logits with the current prototype bank:
$q_{i,g}^{a}=(z_i^a)^\top c_g^a/\tau,\ g=1,\ldots,G$.

\STATE Construct target label $t_i$: one-hot target for uncensored patient ($c_i=0$) and soft target over candidate risk groups for censored patient ($c_i=1$).

\STATE Compute prototype-based cross-entropy loss:
$\mathcal{L}_{\mathrm{rd}}^{a}=-\sum_{g=1}^{G} t_{i,g}\log\frac{\exp(q_{i,g}^{a})}{\sum_{r=1}^{G}\exp(q_{i,r}^{a})}$.


\IF{$c_i=0$}
    \STATE Update the assigned risk prototype:
    $c_{y_i}^{a}\leftarrow \operatorname{Norm}(\mu c_{y_i}^{a}+(1-\mu)z_i^a)$.
\ELSE
    \STATE Update the assigned and lower-risk candidate prototypes with smaller scale:
    $\mu_c=1-(1-\mu)\lambda_u$.
    \FOR{$g=y_i,\ldots,G$}
        \STATE $c_g^{a}\leftarrow \operatorname{Norm}(\mu_c c_g^{a}+(1-\mu_c)z_i^a)$.
    \ENDFOR
\ENDIF

\RETURN $\mathcal{L}_{\mathrm{rd}}^{a}$ and updated $C^a$
\end{algorithmic}
\end{algorithm}

Although RP Compression and multi-branch Q-Former learn compact multimodal features, high-dimensional inputs may still cause the model to capture task-irrelevant interactions, weakening discrimination and generalization. 
To address this issue, we propose patient-level Risk-discriminative Learning (RL), which maintains dynamic risk prototypes after the Q-Former to guide the organization of patient representations. By pulling patient representations within the same risk group closer to their shared prototype and separating representations from different risk groups, RL encourages the model to focus on patient-level risk patterns rather than incidental correlations.

The computation process is shown in Algorithm~\ref{alg:risk_discriminated_learning}. For each training step, the algorithm operates on one patient sample $i$ and one modality $a\in\{\mathrm{wsi},\mathrm{omic}\}$. Here, $G$ denotes the number of discretized risk groups (default is 4). The main hyperparameters include the temperature $\tau$ for prototype logits, the Exponential Moving Average (EMA)~\cite{tarvainen2017ema} momentum $\mu$ for prototype updates, and the censored update scale $\lambda_u$. The risk-discriminative loss is $\mathcal{L}_{\mathrm{rd}}=\mathcal{L}_{\mathrm{rd}}^{\mathrm{wsi}}+\mathcal{L}_{\mathrm{rd}}^{\mathrm{omic}}$. Consequently, the overall training loss can be expressed as $\mathcal{L}=\mathcal{L}_{\mathrm{surv}}+\beta \mathcal{L}_{\mathrm{rd}}$, where $\mathcal{L}_{\mathrm{surv}}$ denotes survival prediction loss~\cite{zhou2024cohort}. This design is more efficient than clustering-based prototype construction, since the risk prototypes are updated online through EMA without iterative clustering. Meanwhile, EMA enables the prototypes to continuously track the feature distribution as model parameters are updated during training. Furthermore, the prototype bank stores one vector for each risk group, requiring only four prototype vectors in each modality and thus introducing negligible memory overhead.

\section{Experiments}

\begin{table*}[t]
\centering
\renewcommand{\arraystretch}{1.12}
\setlength{\tabcolsep}{4.5pt}
\resizebox{\textwidth}{!}{
\begin{tabular}{l|c|ccccccc}
\hline
\textbf{Model}
& \textbf{Modality}
& \textbf{BLCA}
& \textbf{BRCA}
& \textbf{GBMLGG}
& \textbf{LUAD}
& \textbf{UCEC}
& \textbf{LIHC}
& \textbf{Avg} \\
\hline

ABMIL$_{\text{path}}$
& t.
& $0.600_{\pm 0.028}$ & $0.637_{\pm 0.043}$ & $0.767_{\pm 0.039}$
& $0.613_{\pm 0.026}$ & $0.636_{\pm 0.041}$
& $0.638_{\pm 0.024}$
& $0.649$ \\

ABMIL$_{\text{gene}}$
& t.
& $0.542_{\pm 0.029}$ & $0.595_{\pm 0.051}$ & $0.798_{\pm 0.030}$
& $0.563_{\pm 0.032}$ & $0.641_{\pm 0.056}$
& $0.652_{\pm 0.059}$
& $0.632$ \\

\hline

SNN
& g.
& $0.643_{\pm 0.040}$ & $0.638_{\pm 0.039}$ & $0.832_{\pm 0.015}$
& $0.652_{\pm 0.036}$ & $0.660_{\pm 0.034}$
& $0.655_{\pm 0.039}$
& $0.680$ \\

SNNTrans
& g.
& $0.659_{\pm 0.033}$ & $0.658_{\pm 0.054}$ & $0.839_{\pm 0.006}$
& $0.653_{\pm 0.031}$ & $0.648_{\pm 0.018}$
& $0.663_{\pm 0.033}$
& $0.687$ \\

\hline

MaxMIL
& p.
& $0.620_{\pm 0.054}$ & $0.649_{\pm 0.057}$ & $0.777_{\pm 0.026}$
& $0.643_{\pm 0.039}$ & $0.613_{\pm 0.034}$
& $0.654_{\pm 0.024}$
& $0.659$ \\

MeanMIL
& p.
& $0.637_{\pm 0.033}$ & $0.638_{\pm 0.032}$ & $0.810_{\pm 0.024}$
& $0.634_{\pm 0.032}$ & $0.627_{\pm 0.036}$
& $0.662_{\pm 0.054}$
& $0.668$ \\

ABMIL
& p.
& $0.638_{\pm 0.011}$ & $0.654_{\pm 0.029}$ & $0.810_{\pm 0.023}$
& $0.631_{\pm 0.044}$ & $0.660_{\pm 0.032}$
& $0.663_{\pm 0.037}$
& $0.676$ \\

CLAM
& p.
& $0.633_{\pm 0.022}$ & $0.659_{\pm 0.065}$ & $0.802_{\pm 0.022}$
& $0.626_{\pm 0.038}$ & $0.656_{\pm 0.020}$
& $0.668_{\pm 0.038}$
& $0.674$ \\

TransMIL
& p.
& $0.630_{\pm 0.026}$ & $0.669_{\pm 0.025}$ & $0.810_{\pm 0.030}$
& $0.624_{\pm 0.043}$ & $0.660_{\pm 0.039}$
& $0.657_{\pm 0.034}$
& $0.675$ \\

\hline

MCAT
& p.+g.
& $0.674_{\pm 0.025}$ & $0.665_{\pm 0.032}$ & $0.840_{\pm 0.028}$
& $0.667_{\pm 0.033}$ & $0.672_{\pm 0.034}$
& $0.671_{\pm 0.028}$
& $0.698$ \\

CMTA
& p.+g.
& $0.684_{\pm 0.035}$ & $0.661_{\pm 0.025}$ & $0.846_{\pm 0.022}$
& $0.683_{\pm 0.039}$ & $0.686_{\pm 0.035}$
& $0.679_{\pm 0.041}$
& $0.707$ \\

MOTCat
& p.+g.
& $0.672_{\pm 0.028}$ & $0.675_{\pm 0.016}$ & $0.853_{\pm 0.031}$
& $0.678_{\pm 0.023}$ & $0.681_{\pm 0.031}$
& $0.670_{\pm 0.025}$
& $0.705$ \\

CCL
& p.+g.
& $0.650_{\pm 0.039}$ & $0.678_{\pm 0.058}$ & $0.847_{\pm 0.018}$
& $0.690_{\pm 0.021}$ & $0.692_{\pm 0.045}$
& $0.680_{\pm 0.044}$
& $0.706$ \\

MoME
& p.+g.
& $0.685_{\pm 0.028}$ & $0.688_{\pm 0.026}$ & $0.846_{\pm 0.014}$
& $0.694_{\pm 0.029}$ & $0.686_{\pm 0.024}$
& $0.685_{\pm 0.042}$
& $0.714$ \\

LD-CVAE
& p.+g.
& $0.679_{\pm 0.033}$ & $0.673_{\pm 0.037}$ & $0.852_{\pm 0.027}$
& $0.682_{\pm 0.034}$ & $0.690_{\pm 0.035}$
& $0.684_{\pm 0.041}$
& $0.710$ \\

DeReF
& p.+g.
& $0.687_{\pm 0.015}$ & $0.679_{\pm 0.035}$ & $0.850_{\pm 0.013}$
& $0.678_{\pm 0.021}$ & $0.693_{\pm 0.041}$
& $0.681_{\pm 0.026}$
& $0.711$ \\

HDMoE
& p.+g.
& $0.698_{\pm 0.036}$ & $0.694_{\pm 0.032}$ & $0.853_{\pm 0.025}$
& $0.683_{\pm 0.025}$ & $0.696_{\pm 0.025}$
& $0.683_{\pm 0.021}$
& $0.718$ \\
\hline

PS3
& t.+p.+g.
& $0.691_{\pm 0.023}$ & $0.688_{\pm 0.036}$ & $0.851_{\pm 0.013}$
& $0.680_{\pm 0.031}$ & $0.691_{\pm 0.019}$
& $0.689_{\pm 0.027}$
& $0.715$ \\

\rowcolor{gray!15} 
\textbf{CRMSurv}
& \textbf{t.+p.+g.}
& $\mathbf{0.704_{\pm 0.023}}$
& $\mathbf{0.702_{\pm 0.019}}$
& $\mathbf{0.856_{\pm 0.017}}$
& $\mathbf{0.703_{\pm 0.020}}$
& $\mathbf{0.712_{\pm 0.022}}$
& $\mathbf{0.693_{\pm 0.025}}$
& $\mathbf{0.728}$ \\

\hline
\end{tabular}
}
\caption{C-Index (mean $_{\pm}$ std) performance over six TCGA datasets. The best results are highlighted in bold. Here, t., p., and g. denote Textual, Pathology, and Genomic modalities, respectively.}
\label{tab:main_results}
\end{table*}

\subsection{Datasets}
We evaluate our proposed CRMSurv using publicly available datasets from The Cancer Genome Atlas (TCGA\footnote{https://portal.gdc.cancer.gov}) across six types of cancer: Bladder Urothelial Carcinoma (BLCA) (n=342), Breast Invasive Carcinoma (BRCA) (n=900), Glioblastoma Lower Grade Glioma (GBMLGG) (n=512), Lung Adenocarcinoma (LUAD) (n=429), Uterine Corpus Endometrial Carcinoma (UCEC) (n=479) and Liver Hepatocellular Carcinoma (LIHC) (n=304). These data encompass hundreds of paired genomic and WSI data, along with follow-up data from multiple centers. The WSIs are presented by HE staining. For genomic data, consistent with prior studies~\cite{CMTAs}, we use RNA-seq, CNV, and Mutation data and further group them into six genomic sub-sequences: Tumor Suppression, Oncogenesis, Protein Kinases, Cellular Differentiation, Transcription, and Cytokines and Growth. Furthermore, we refine and construct WSI reports~\cite{kefeli2024tcga} and Genomic reports (Details are provided in Appendix) as text modality. 

\subsection{Implementation Details}
\textbf{Training Settings.} We compare CRMSurv against a wide range of unimodal and multimodal baselines. For unimodal baselines, we use ABMIL~\cite{AttnMIL} for modeling pathology and genomic report embeddings, SNN and SNNTrans~\cite{SNN} for genomic data, and MaxMIL, MeanMIL, ABMIL~\cite{AttnMIL}, CLAM~\cite{CLAM}, and TransMIL~\cite{TransMIL} for WSI modeling. For multimodal baselines, we include MCAT~\cite{MCAT}, CMTA~\cite{CMTAs}, MOTCAT~\cite{MOTCat}, CCL~\cite{zhou2024cohort}, MoME~\cite{xiong2024mome}, LD-CVAE~\cite{LD-CVAE}, DeReF~\cite{wang2025decouple}, HDMoE~\cite{wang2026hdmoe} and PS3~\cite{Raza2025PS3AM}.

\textbf{Evaluation.} We conduct five-fold cross-validation and utilize Concordance Index (C-Index)~\cite{harrell1982evaluating} to evaluate the performance of correctly ranking overall survival based on predicted patient risk scores. Additionally, Kaplan-Meier~\cite{kaplan1958nonparametric} and T-test analysis are used to assess the statistical significance of differences in survival prediction between high-risk and low-risk groups.

\textbf{Hyperparameters.} 
Our framework is implemented in PyTorch and trained on a single NVIDIA GeForce RTX 3090 GPU. 
We train the model for 30 epochs using AdamW with an initial learning rate of $5\times10^{-4}$ and weight decay of $1\times10^{-5}$. The hidden dimension is $d=256$. For RP compression, we retain $25\%$ of patches, i.e., $k=\lfloor n/4 \rfloor$, and set the prototype number to $r=8$. The query number is $h=8$, and the loss weight is $\beta=0.1$. In RL, we set $\tau=0.2$ and $\mu=0.5$. For censored samples, the update scale $\lambda_u$ is averaged over number of the candidate risk groups. 


\subsection{Main Results}
Results on six TCGA datasets are shown in Table~\ref{tab:main_results}. Overall, CRMSurv achieves the best performance across all datasets, with an average C-Index of $0.728$.

For unimodal methods, the text-based ABMIL$_{\text{path}}$ and ABMIL$_{\text{gene}}$ achieve C-Index values above $0.5$ across all datasets, indicating that WSI and genomic reports contain useful semantic information for survival prediction and can serve as complementary modalities. Compared with textual modalities, genomic numerical features show more stable performance. For example, SNNTrans achieves an average C-Index of $0.687$, outperforming most unimodal WSI methods and highlighting the prognostic value of genomic data. Meanwhile, with the pathology-specific CONCH encoder, several WSI-based methods achieve results comparable to or better than genomic models on specific datasets. For instance, TransMIL reaches $0.660$ on UCEC, surpassing SNNTrans with $0.648$, suggesting that high-quality WSI features also provide important prognostic information.

Compared with unimodal methods, multimodal methods generally achieve higher C-Index values, demonstrating that interactions between pathology and genomics capture richer associations and complementary information. However, most multimodal baselines mainly fuse pathology and genomic features, with limited use of report-derived textual semantics. PS3 introduces pathology report semantics and achieves an average C-Index of $0.715$, higher than many bimodal fusion methods, indicating the value of textual semantics. Nevertheless, PS3 only uses pathology report semantics, leaving genomic report semantics unexplored. Moreover, its prototype-based WSI representation may over-compress local pathology details and cause information loss.
In contrast, CRMSurv leverages textual semantics from both pathology and genomic reports. Through the RP Compression module, CRMSurv reduces patch numbers while preserving local pathology information. Quantitative results show that CRMSurv achieves the best performance on all six datasets, with an average C-Index of $0.728$, improving over PS3 by $1.3\%$. In addition, its average standard deviation over five-fold cross-validation is only $0.021$, indicating more stable performance across data splits. These results demonstrate that patient-level RL helps model learn more stable prognostic representations and improves its generalization.

\subsection{Ablation Studies}
\begin{figure*}[t]
    \centering
    \includegraphics[width=\linewidth]{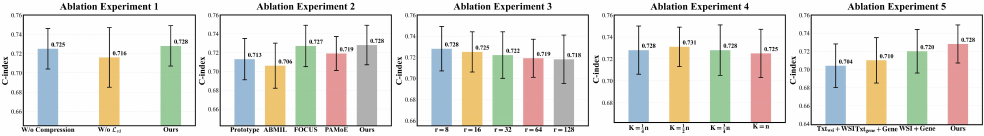}
    \caption{Five ablation studies across six TCGA datasets. Each result reports the C-index and standard deviation averaged over the six datasets.}
    \label{fig:ablation}
\end{figure*}

We conduct five ablation studies to validate the effectiveness of CRMSurv, with the results reported in Figure~\ref{fig:ablation}.



\begin{figure*}[t]
    \centering
    \includegraphics[width=\linewidth]{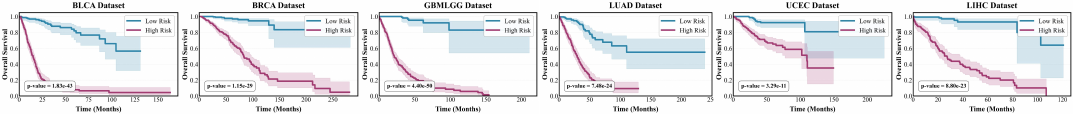}
    \caption{Visualization of Kaplan-Meier Analysis on six TCGA datasets, where patient stratifications of low risk (blue) and high risk (red) are presented. Shaded areas refer to the confidence intervals.}
    \label{fig:surv}
\end{figure*}
\begin{figure}
    \centering
    \includegraphics[width=\linewidth]{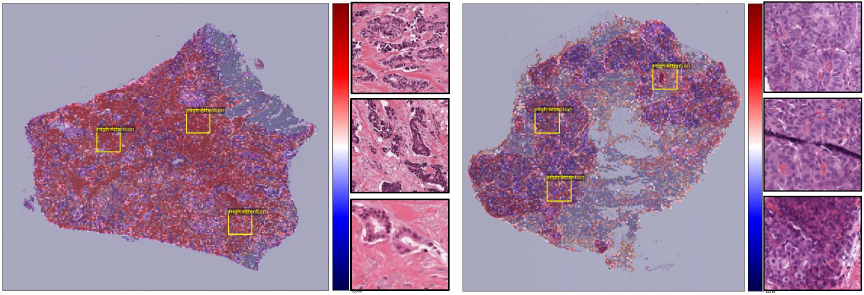}
    \caption{Attention Visualization of two BRCA samples.}
    \label{fig:wsi_vis}
\end{figure}




\textbf{Experiment 1: Impact of Framework Components.}
We evaluate the contribution of key components in CRMSurv. Removing RP Compression yields comparable performance ($\downarrow 0.3\%$), indicating that it can greatly reduce WSI patch numbers while minimizing information loss and preserving effective pathology cues. 
In contrast, removing $\mathcal{L}_{\mathrm{rd}}$ causes performance degradation and larger standard deviation ($0.716$ vs. $0.728$), showing that RL helps the model identify prognosis-relevant features rather than incidental correlations, thereby improving stability and generalization.

\textbf{Experiment 2: Impact of RP Compression.}
To further validate the RP Compression, we replace it with existing WSI patch compression methods while keeping other modules unchanged. We compare attention-based compression (ABMIL~\cite{AttnMIL}), prototype-based compression~\cite{zhou2024cohort}, similarity-based compression (FOCUS~\cite{guo2025focus}), and textual-semantic-based compression (PAMoE~\cite{PAMoE}). ABMIL and prototype-based compression show lower performance ($\downarrow 2.2\%$, $\downarrow 1.5\%$), suggesting that aggregating all patches into few representations may lose local details. Although FOCUS and PAMoE achieve comparable results, they usually retain over half of the WSI patches. In contrast, RP Compression obtains competitive performance with fewer patches, indicating a better balance between compression efficiency and information preservation.

\textbf{Experiment 3: Impact of Cluster Number $r$.}
We evaluate the effect of the prototype number $r$ in RP Compression. $r=8$ achieves the best average C-Index ($0.728$), while performance gradually decreases as more prototypes are used. This suggests that a compact prototype set provides a proper reconstructability reference. Specifically, with fewer prototypes, reconstruction mainly captures dominant morphological patterns, so patches carrying additional local information tend to produce larger and more distinguishable errors. In contrast, too many prototypes make reconstruction overly fine-grained, narrowing the error gap among patches and weakening the ability to highlight truly informative patches. Moreover, excessive prototype tokens may increase the burden of Q-Former to select useful information. Therefore, $r=8$ achieves a better balance between informative patch selection and prototype-based information compensation.

\textbf{Experiment 4: Impact of Compression Ratio $K$.}
To analyze the influence of retained patch numbers, we compare different compression ratios. The results show that retaining $k=\frac{1}{4}n$ and $k=\frac{1}{2}n$ WSI patches achieves similar and better performance ($0.728$ vs. $0.731$), indicating that a smaller subset of high-value patches is sufficient for effective survival prediction. In contrast, retaining $k=\frac{3}{4}n$ or all patches slightly decreases performance, suggesting that more patches do not necessarily provide more useful prognostic information and may dilute the contribution of key local patterns. Considering both predictive performance and computational efficiency, we set the compression ratio $k=\frac{1}{4}n$ by default.

\textbf{Experiment 5: Impact of Modality Importance.}
We evaluate different modality combinations. The results show that combining report-derived textual semantics with either WSI ($0.704$) or genomic features ($0.710$) yields relatively limited performance. In contrast, combining WSI and genomic profiles improves the C-Index to $0.720$, confirming the complementarity between pathology and molecular information. Further introducing report-derived textual semantics increases performance to $0.728$, indicating that textual reports provide additional prognostic information beyond pathology and genomic features. These results show that jointly modeling pathology, genomics, and report semantics yields more comprehensive patient representations.


In addition, we conduct more experiments on $\beta$ sensitivity, the number of learnable queries, etc., as detailed in Appendix.


\subsection{Visualization Analysis}

\textbf{Kaplan-Meier Analysis.} We use Kaplan-Meier analysis~\cite{kaplan1958nonparametric} to evaluate the risk stratification ability of CRMSurv. Patients are divided into high- and low-risk groups according to the median predicted risk score, and the log-rank test is used to assess the difference between the two groups. As shown in Figure~\ref{fig:surv}, CRMSurv achieves log-rank $p$-values below $0.05$ on all datasets, indicating significant separation between high- and low-risk patients.

\textbf{Attention Analysis of WSI.} To validate the model's ability to identify key informative regions, we select representative low- (left) and high- (right) risk cases from the BRCA dataset and visualize the attention between compressed WSI features after Q-Former and retained original patches, as shown in Figure~\ref{fig:wsi_vis}. From the enlarged patches, the attended regions contain evident tumor-related structures and show some local differences between low- and high-risk cases, such as glandular-like patterns with limited disordered cell clusters in low-risk regions, and broader dense cellular arrangements with stronger atypia in high-risk regions. These observations demonstrate the effectiveness of the proposed RP compression and fusion method in preserving and querying information associated with prognostic risk.


We also visualize prototypes and retained patches to assess their morphological complementarity, examine the impact of $\mathcal{L}_{\mathrm{rd}}$ on feature distributions, and perform T-tests, SHAP analyses, etc., as detailed in Appendix.

\section{Conclusions}

In this paper, we proposed CRMSurv for obtaining a set of compact, information-preserving, and prognosis-oriented representations. We developed RP Compression to combine compact prototype summaries with high-reconstruction-error patches, preserving global morphology and fine-grained local evidence with fewer WSI tokens. We further introduced patient-level RL to organize representations around stable prognostic patterns and reduce overfitting to irrelevant information. Finally, we employed multi-branch Q-Formers and cross-attention to adapt and align the refined modality representations. Experiments on six cancer datasets demonstrated a 1\% average improvement over the strongest baseline while retaining only 25\% of WSI patches.

\bibliography{AnonymousSubmission2027}
\clearpage
\section{Appendix}
\subsection{Multi-branch Q-Former Block}
\begin{figure}[h]
    \centering
    \includegraphics[width=\linewidth]{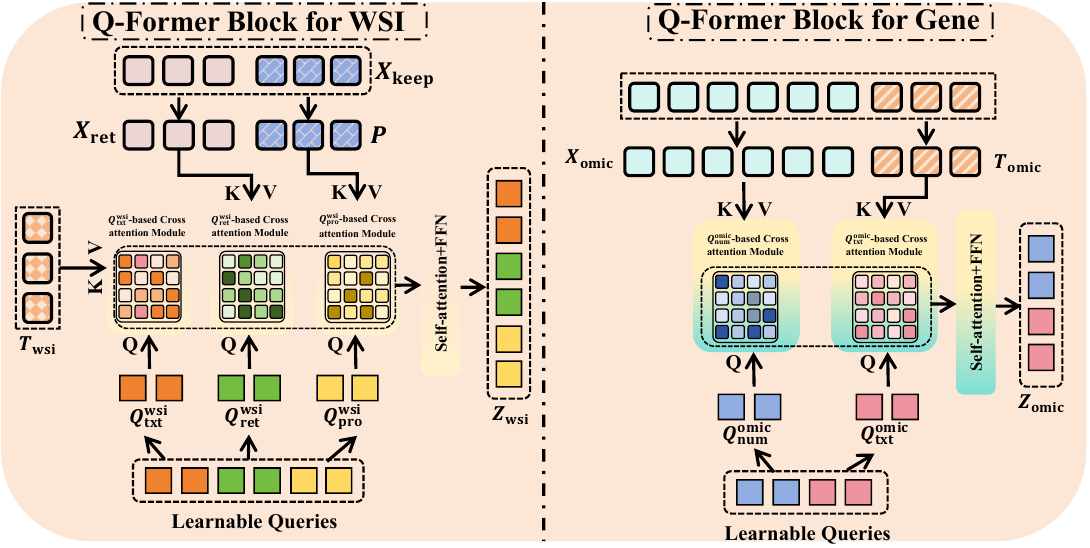}
    \caption{Multi-branch Q-Former for two modalities.}
    \label{fig:q_former}
\end{figure}

The Multi-branch Q-Former Block is shown in Figure~\ref{fig:q_former}. Taking the genomic modality as an example, our multi-branch Q-Former initializes learnable query tokens 
$Q^{\mathrm{omic}}\in \mathbb{R}^{2h\times d}$ and divides them into two equal branches:
$Q^{\mathrm{omic}}=[Q^{\mathrm{omic}}_{\mathrm{num}};Q^{\mathrm{omic}}_{\mathrm{txt}}]$.
The numerical branch extracts molecular patterns from genomic numerical features, while the textual branch extracts mechanism-level semantics from genomic report features. 
For a query group $Q$ and input tokens $X$, the cross-attention operation is defined as:
\begin{equation}
    \operatorname{CrossAttn}(Q,X,X)
    =
    \operatorname{Softmax}
    \left(
    \frac{(QW_q)(XW_k)^{\top}}{\sqrt{d}}
    \right)
    XW_v ,
\end{equation}
where $W_q$, $W_k$, and $W_v$ are learnable projection matrices.

Based on this operation, the two genomic branches are computed as:
\begin{equation}
\begin{aligned}
\hat{Q}^{\mathrm{omic}}_{\mathrm{num}}
&=
\operatorname{CrossAttn}
(Q^{\mathrm{omic}}_{\mathrm{num}},X_{\mathrm{omic}},X_{\mathrm{omic}}),\\
\hat{Q}^{\mathrm{omic}}_{\mathrm{txt}}
&=
\operatorname{CrossAttn}
(Q^{\mathrm{omic}}_{\mathrm{txt}},T_{\mathrm{omic}},T_{\mathrm{omic}}).
\end{aligned}
\end{equation}
Then, the two query groups are concatenated and integrated by self-attention and a feed-forward network:
\begin{equation}
    Z_{\mathrm{omic}}
    =
    \operatorname{FFN}
    \left(
    \operatorname{SelfAttn}
    \left(
    [\hat{Q}^{\mathrm{omic}}_{\mathrm{num}};
    \hat{Q}^{\mathrm{omic}}_{\mathrm{txt}}]
    \right)
    \right).
\end{equation}
In this way, the genomic representation jointly encodes pathway-level molecular patterns and report-level mechanism semantics in a unified query space.

The pathology modality follows the same multi-branch design, but uses three branches for retained WSI tokens, compact prototype tokens, and WSI report features. 
Through this multi-branch query design, each modality first performs internal numerical-semantic adaptation and information selection, avoiding premature fusion of heterogeneous features before proper alignment.

\subsection{Refinement of the WSI Report}
To better utilize WSI report semantics and reduce redundant or unstructured descriptions, we design an LLM-based WSI report refinement strategy. Specifically, for each patient, we construct a structured prompt and feed it into Qwen2.5-32B-Instruct to extract compact prognosis-relevant pathology textual features.
The prompt contains three types of information. First, patient clinical information, including cancer type, case ID, age, sex, censorship status, survival time, pathology TNM stage, pathology stage, and clinical stage, provides patient-level context. Second, the raw pathology report, which provides patient-specific diagnostic descriptions. Third, the cancer-specific pathology concepts, which are used as background guidance. Following the prior work~\cite{sun2024pathgen}, we use the 20 pre-constructed cancer-specific pathology concepts provided in that work, covering common tumor-related manifestations such as nuclear atypia, mitotic activity, necrosis, invasive growth, lymphovascular invasion, differentiation status, and inflammatory infiltration. These concepts only guide the LLM to focus on potentially prognosis-relevant pathology cues, rather than serving as hard selection rules.

\begin{figure*}
    \centering
    \includegraphics[width=\linewidth]{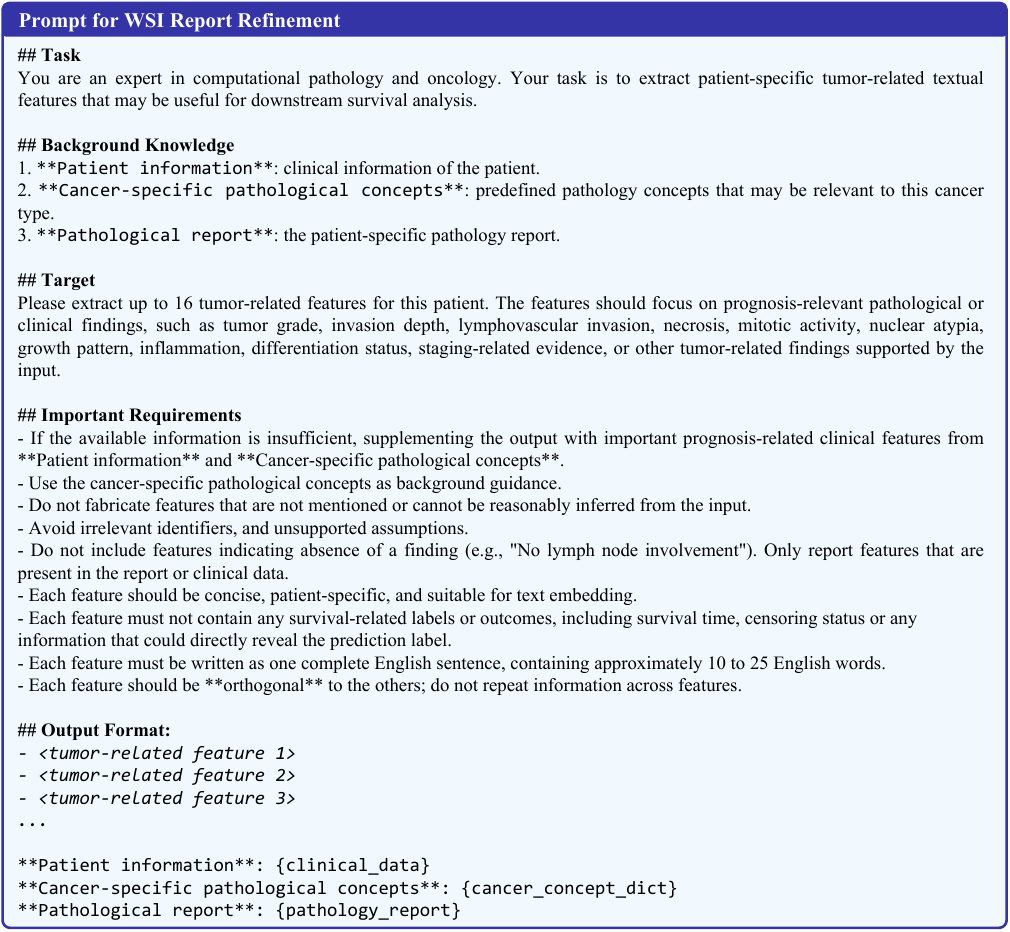}
    \caption{The prompt for WSI report refinement.}
    \label{fig:wsirefine}
\end{figure*}

\begin{figure*}[!t]
    \centering
    \includegraphics[width=\linewidth]{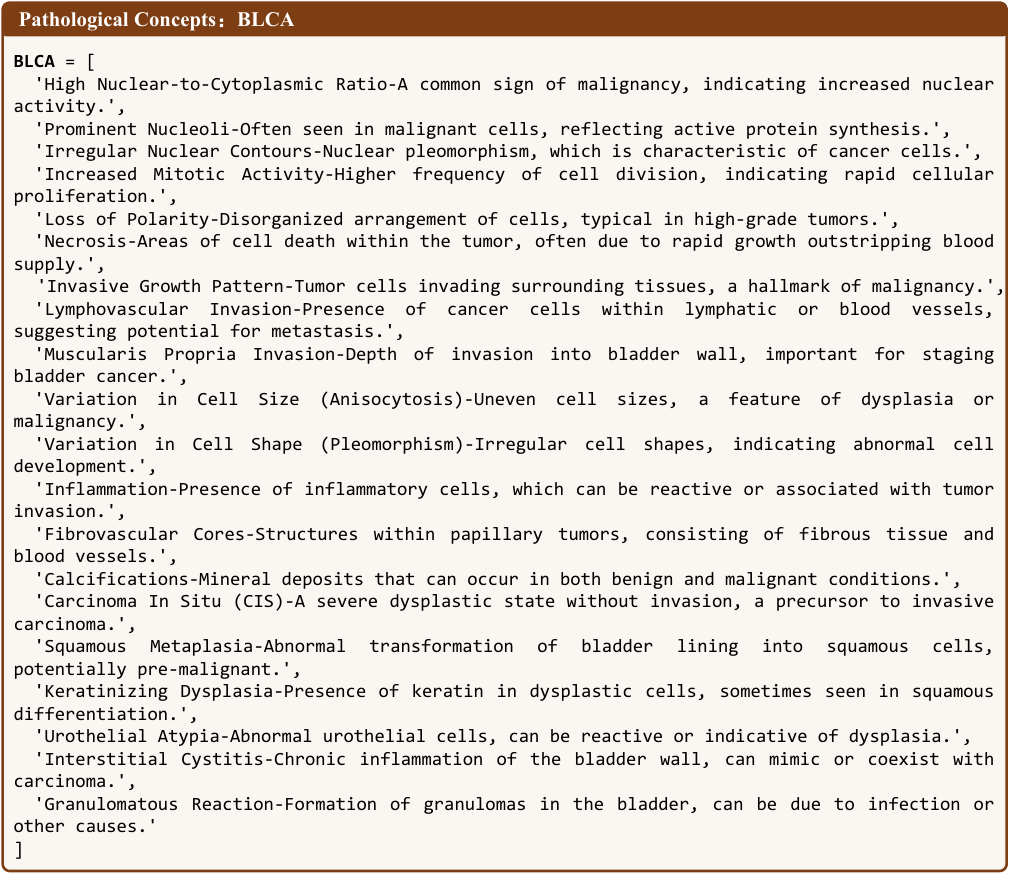}
    \caption{Illustration of twenty pathology concepts about BLCA.}
    \label{fig:wsiconcept}
\end{figure*}

For the output, we require the LLM to extract 16 patient-specific tumor-related textual features using only the available clinical and pathology information, without exposing censoring status or survival labels.
Each feature is a concise and complete English sentence focusing on prognosis-relevant findings, such as tumor grade, invasion depth, necrosis, mitotic activity, nuclear atypia, growth pattern, inflammatory response, differentiation status, and staging evidence. To ensure reliability and suitability for text embedding, the prompt prevents fabricated information, irrelevant identifiers, repetitive descriptions, and negative findings. Finally, each raw WSI report is converted into a compact, non-redundant, and prognosis-relevant set of textual semantic features for subsequent multimodal survival analysis. The detailed prompt is shown in Figure~\ref{fig:wsirefine} and Figure~\ref{fig:wsiconcept}.

\subsection{Construction of the Genomic Report}
\begin{figure*}[t]
    \centering
    \includegraphics[width=\linewidth]{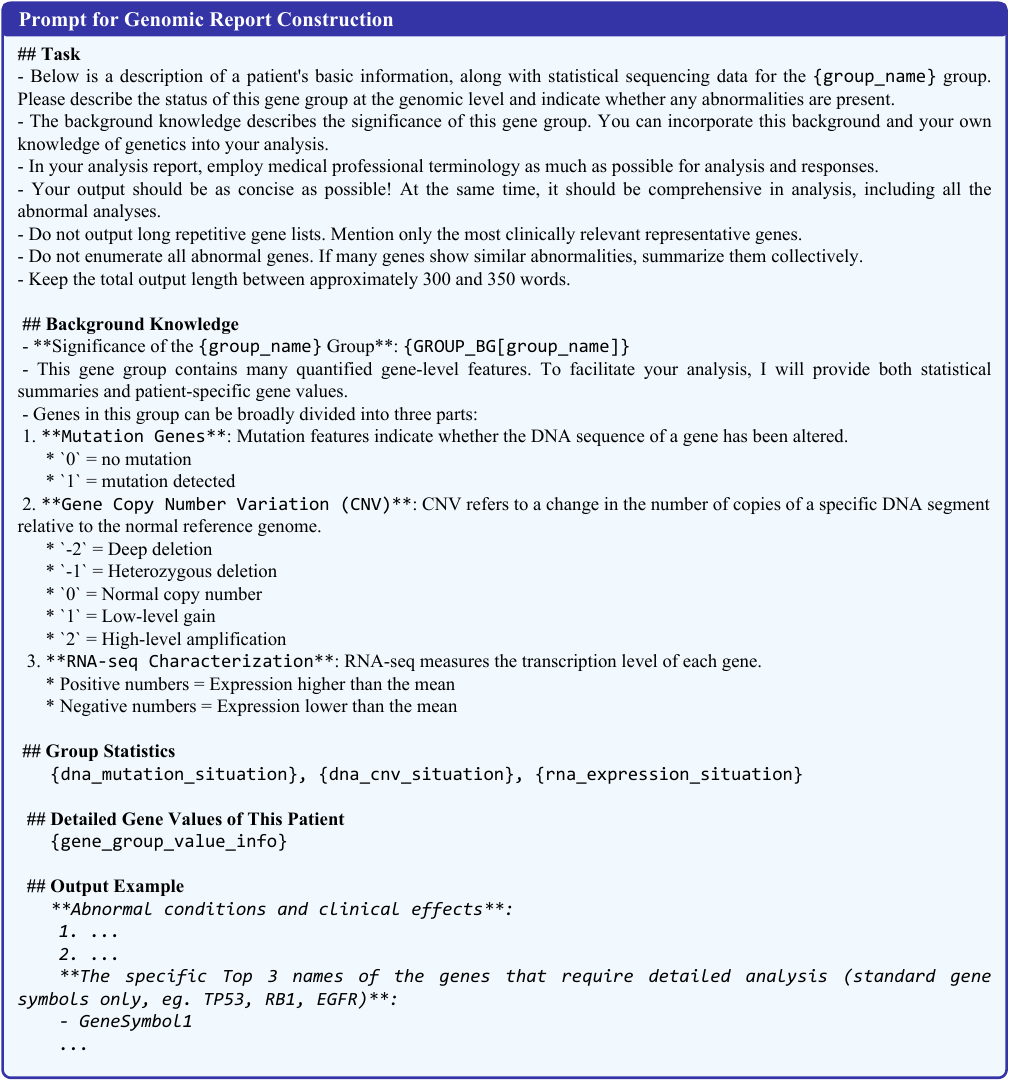}
    \caption{The prompt for Genomic report construction.}
    \label{fig:geneconstruct}
\end{figure*}

To construct patient-level genomic textual reports, we follow the analysis paradigm of SurvAgent~\cite{huang2025survagent} and divide genes into six functional categories: tumor suppressor genes, oncogenes, protein kinases, cell differentiation markers, transcription factors, and cytokines and growth factors. This categorization organizes high-dimensional genomic data into biologically meaningful functional subspaces. For each patient, we first compute abnormality statistics within each gene category, such as mean expression, median expression, and mutation ratio, to summarize category-level molecular alterations. We then construct a prompt using the gene numerical information and statistical descriptions, as shown in Figure~\ref{fig:geneconstruct}, and input it into Qwen2.5-32B-Instruct to summarize potentially important abnormal signals. Finally, several important genes identified in the summaries are further annotated using MyGene~\cite{wu2013mygene}, and the retrieved descriptions are appended to the generated reports.

\begin{figure*}[t]
    \centering
    \includegraphics[width=\linewidth]{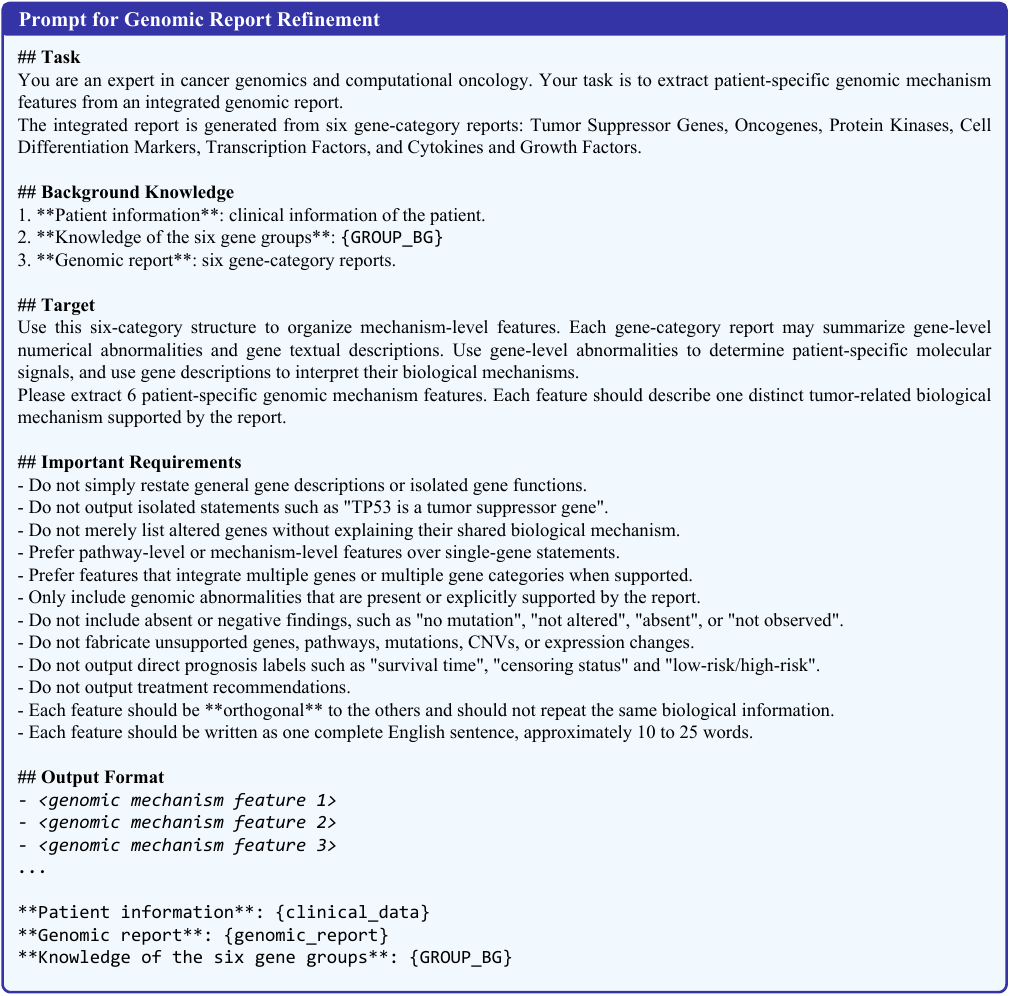}
    \caption{The prompt for Genomic report refinement.}
    \label{fig:generefine}
\end{figure*}

\begin{figure*}[!t]
    \centering
    \includegraphics[width=\linewidth]{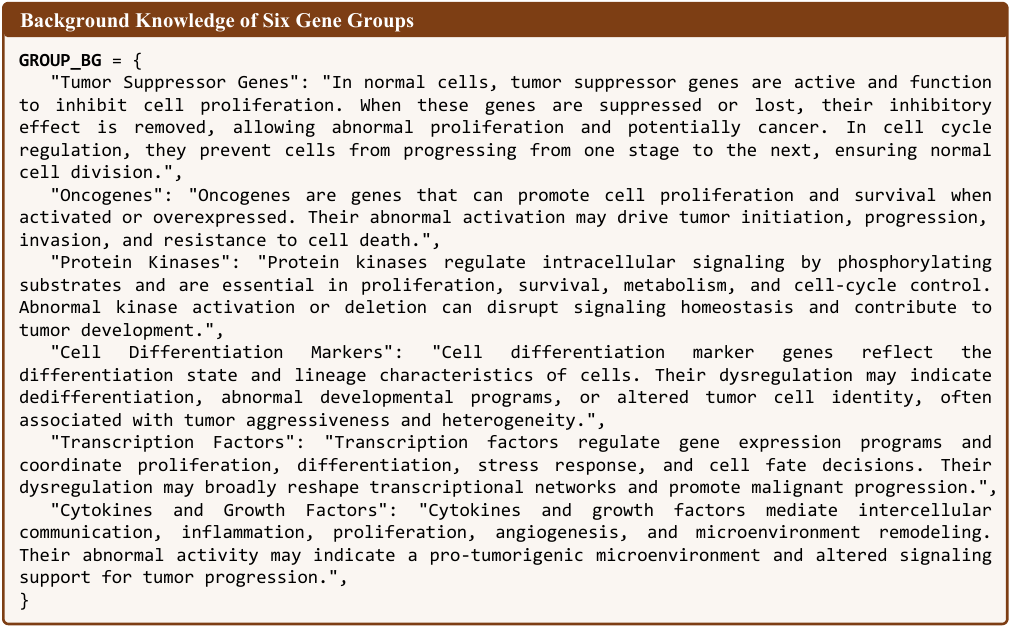}
    \caption{Illustration of functional interpretation backgrounds for the six gene categories.}
    \label{fig:6genebg}
\end{figure*}

Based on these reports, we further design a genomic report refinement strategy to obtain compact, mechanism-oriented, and embedding-friendly genomic semantic representations. Specifically, we construct a structured refinement prompt, as shown in Figure~\ref{fig:generefine} and Figure~\ref{fig:6genebg}, and input it into Qwen2.5-32B-Instruct. The prompt contains patient clinical information, gene-category reports, and functional backgrounds for the six gene categories. The clinical information includes cancer type, case ID, age, sex, censoring status, and survival time. The gene-category reports provide patient-specific genomic abnormalities, while the functional backgrounds guide the LLM to interpret abnormal signals from category-level biological perspectives.

We require Qwen2.5-32B-Instruct to extract six patient-specific genomic mechanism features from the gene-category reports. Each feature describes a tumor-related molecular mechanism supported by the patient’s genomic abnormalities, rather than simply restating individual gene functions or listing altered genes. To ensure reliability and suitability for text embedding, the prompt constrains the LLM to use only explicitly supported genomic abnormalities and avoid unsupported inferences, direct prognostic labels, and treatment recommendations. It also encourages semantic non-redundancy across features. Through this process, the six gene-category reports are refined into compact, mechanism-level, and patient-specific genomic semantic representations for subsequent multimodal modeling.

\subsection{Data Preparation}

We use publicly available datasets from TCGA as input, including WSI images, genomic profiles, and WSI reports. In addition, we construct genomic reports to provide high-level molecular semantic information complementary to genomic numerical features.

\textbf{WSI Images.}
Following the common setting~\cite{CMTAs,MCAT}, we use CLAM~\cite{CLAM} to divide each WSI into non-overlapping $256 \times 256$ patches at a $20\times$ magnification level. We then employ CONCH~\cite{lu2024conch}, a pathology-specific feature extractor, to obtain patch-level features, denoted as
$\tilde{X}_{\mathrm{wsi}}=\{x^w_1,x^w_2,\ldots,x^w_n\}\in \mathbb{R}^{n\times d_w}$, where $n$ and $d_w$ represent the patch number and feature dimension, respectively.

\textbf{WSI Reports.}
We obtain curated WSI reports from~\cite{kefeli2024tcga}. An LLM is then used to extract $m$ prognosis-relevant pathology descriptions from each report. The detailed refinement process is provided in the above section.
Finally, we use the pretrained text encoder of CONCH~\cite{lu2024conch} to extract report features:
$\tilde{T}_{\mathrm{wsi}}=\{t^w_1,t^w_2,\ldots,t^w_m\}\in \mathbb{R}^{m\times d_t}$.

\textbf{Genomic Profiles.}
Following previous studies~\cite{CMTAs,MCAT}, we extract six genomic sub-sequence features using a Self-Normalizing Neural Network (SNN), producing
$X_{\mathrm{omic}}=\{x^o_1,x^o_2,\ldots,x^o_6\}\in \mathbb{R}^{6\times d}$.

\textbf{Genomic Reports.}
Following~\cite{huang2025survagent}, we summarize numerical abnormalities within six gene categories and use an LLM to generate $l$ patient-specific genomic mechanism descriptions. The detailed construction process is provided in the above section. We then use pretrained PubMedBERT~\cite{Gu2020pubmedbert} to extract genomic report features:
$\tilde{T}_{\mathrm{omic}}=\{t^o_1,t^o_2,\ldots,t^o_l\}\in \mathbb{R}^{l\times d_g}$.

\textbf{Dimension Alignment.}
We set a unified target dimension $d$ for multimodal fusion. After WSI patch compression, $\tilde{X}_{\mathrm{keep}}$ is projected as
$X_{\mathrm{keep}}\in\mathbb{R}^{k\times d}$. The genomic numerical features are directly encoded by SNN into $X_{\mathrm{omic}}\in\mathbb{R}^{6\times d}$. For textual features, two linear projectors align pathology and genomic report features into
$T_{\mathrm{wsi}}\in\mathbb{R}^{m\times d}$ and $T_{\mathrm{omic}}\in\mathbb{R}^{l\times d}$, respectively.

\subsection{Ablation Studies}

\begin{figure*}[t]
    \centering
    \includegraphics[width=\linewidth]{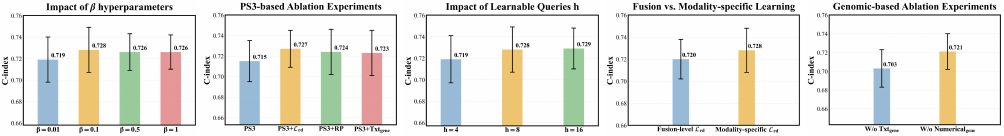}
    \caption{Additional five ablation studies across six TCGA datasets. Each result reports the C-index and standard deviation averaged over the six datasets.}
    \label{fig:ablation_appendix}
\end{figure*}

\begin{table*}[t]
\centering

\resizebox{\textwidth}{!}{
\begin{tabular}{l|ccccccc}
\toprule
\textbf{Method} & \textbf{BLCA} & \textbf{BRCA} & \textbf{GBMLGG} & \textbf{LUAD} & \textbf{UCEC} & \textbf{LIHC} & \textbf{Avg.} \\
\midrule
ReConV
& 0.698$\pm$0.018
& 0.701$\pm$0.022
& \textbf{0.862}$\pm$0.014
& 0.696$\pm$0.021
& 0.704$\pm$0.015
& \textbf{0.702}$\pm$0.016
& 0.727 \\

LRCP
& 0.701$\pm$0.018
& 0.682$\pm$0.019
& 0.842$\pm$0.013
& 0.687$\pm$0.016
& 0.699$\pm$0.017
& 0.674$\pm$0.024
& 0.714 \\

\textbf{RP Comp. (Ours)}
& \textbf{0.704}$\pm$0.023
& \textbf{0.702}$\pm$0.019
& 0.856$\pm$0.017
& \textbf{0.703}$\pm$0.020
& \textbf{0.712}$\pm$0.022
& 0.693$\pm$0.018
& \textbf{0.728} \\
\bottomrule
\end{tabular}
}
\caption{Comparison with reconstruction-based token compression methods. Results are reported as C-index (mean $\pm$ std).}
\label{tab:compression_comparison}
\end{table*}

\begin{figure}
    \centering
    \includegraphics[width=0.6\linewidth]{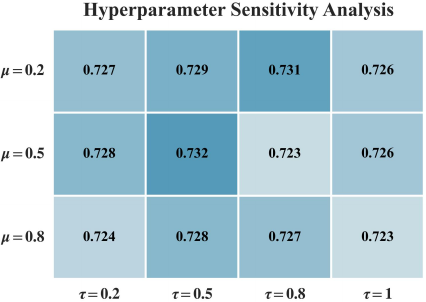}
    \caption{Hyperparameter sensitivity analysis of RL with varying EMA momentum $\mu$ and temperature $\tau$. The result reports the C-index averaged over the six datasets.}
    \label{fig:abl_grid}
\end{figure}


\textbf{Impact of $\beta$ Hyperparameters.} To analyze the sensitivity of the risk-discriminated loss $\mathcal{L}_{\mathrm{rd}}$, we evaluate different settings of the balancing coefficient $\beta$, including $0.01$, $0.1$, $0.5$, and $1$, as shown in Figure~\ref{fig:ablation_appendix}. The results show that the model achieves relatively stable performance when $\beta$ is set between $0.1$ and $1$, indicating that the proposed discrimination constraint is not highly sensitive within a reasonable range. In contrast, when $\beta=0.01$, the performance is relatively weaker, suggesting that an overly small weight cannot provide sufficient risk-discriminative supervision.

\textbf{PS3-based Ablation Experiments.}
To isolate the contributions of proposed individual components, we separately incorporate Risk-discriminative Learning (RL), RP Compression (RP), and genomic reports (Txt$_{gene}$) into PS3, as shown in Figure~\ref{fig:ablation_appendix}. All three components improve performance over the original PS3 (0.715), with RL achieving the largest gain (0.727), followed by RP (0.724) and Txt$_{gene}$ (0.723). These results indicate that risk-discriminative learning contributes most to performance improvements, while RP and genomic reports provide complementary benefits.

\textbf{Impact of Learnable Queries $h$.}
To analyze the effect of learnable query numbers in multi-branch Q-Former, we compare different settings of $h$, as shown in Figure~\ref{fig:ablation_appendix}. The results show that when $h=4$, the model performance slightly decreases ($\downarrow 0.9\%$), suggesting that too few query tokens restrict the representation capacity of the information bottleneck. Increasing $h$ to $8$ or $16$ yields comparable performance ($0.728$ vs. $0.729$). Considering the trade-off between effectiveness and computational cost, we set $h=8$ in our model.

\textbf{Fusion vs. Modality-specific Learning.}
To validate the rationale for applying Q-Former and RL separately to each modality, we compare modality-specific learning with applying them directly to fused representations, as shown in Figure~\ref{fig:ablation_appendix}. The latter leads to a performance drop (0.728 vs. 0.720), suggesting that identifying and adapting key information within each modality before fusion can reduce cross-modal interference and improve survival prediction.

\textbf{Genomic-based Ablation Experiments.}
To evaluate the contributions of different genomic representations, we separately remove genomic numerical features (W/o Numerical$_{gene}$) and textual reports (W/o Txt$_{gene}$), as shown in Figure~\ref{fig:ablation_appendix}. Removing numerical features causes a larger performance drop than removing textual reports (0.703 vs. 0.721), indicating that quantitative molecular information plays a more critical role in survival prediction, while textual semantics provide complementary biological context.

\textbf{Hyperparameter Sensitivity Analysis.}
To evaluate the robustness of RL, we investigate different EMA momentum values $\mu \in \{0.2, 0.5, 0.8\}$ and temperatures $\tau \in \{0.2, 0.5, 0.8, 1.0\}$, as shown in Figure~\ref{fig:abl_grid}. The average C-index remains consistently stable across the tested settings (variation below 0.5\%), demonstrating the robustness of RL to different hyperparameter choices.

\textbf{Comparison with Reconstruction-based Token Compression Methods.}
To evaluate the proposed RP Compression, we compare it with two recent reconstruction-based methods, ReConV~\cite{tang2026reconv} and LRCP~\cite{lu2026lrcp}, as shown in Table~\ref{tab:compression_comparison}. Our method achieves the highest average C-index (0.728), outperforming ReConV (0.727) and LRCP (0.714). Unlike ReConV, which requires additional prototype and reconstruction learning, our method introduces no extra trainable compression parameters. Moreover, LRCP's token merging may dilute fine-grained local information, whereas our method preserves original high-error patches alongside morphological prototypes. This design achieves efficient compression while maintaining feature integrity and inherent interpretability.

\textbf{Computational Efficiency Analysis.}
To evaluate the efficiency of RP Compression, we take TCGA-BLCA cohort as an example and compare CRMSurv with and without patch compression in terms of FLOPs, peak GPU memory, and inference time. By retaining only 25\% of the original patches alongside morphological prototypes, our method reduces average GFLOPs from 893 to 225 ($\downarrow$74.8\%), average peak GPU memory from 4.011 to 1.283 GB ($\downarrow$68.0\%), and average inference time from 0.062 to 0.025 seconds per sample ($\downarrow$59.7\%). These results demonstrate that RP Compression substantially improves computational efficiency while preserving predictive performance.

\subsection{Visualization Analysis}
\begin{figure*}[t]
    \centering
    \includegraphics[width=0.9\linewidth]{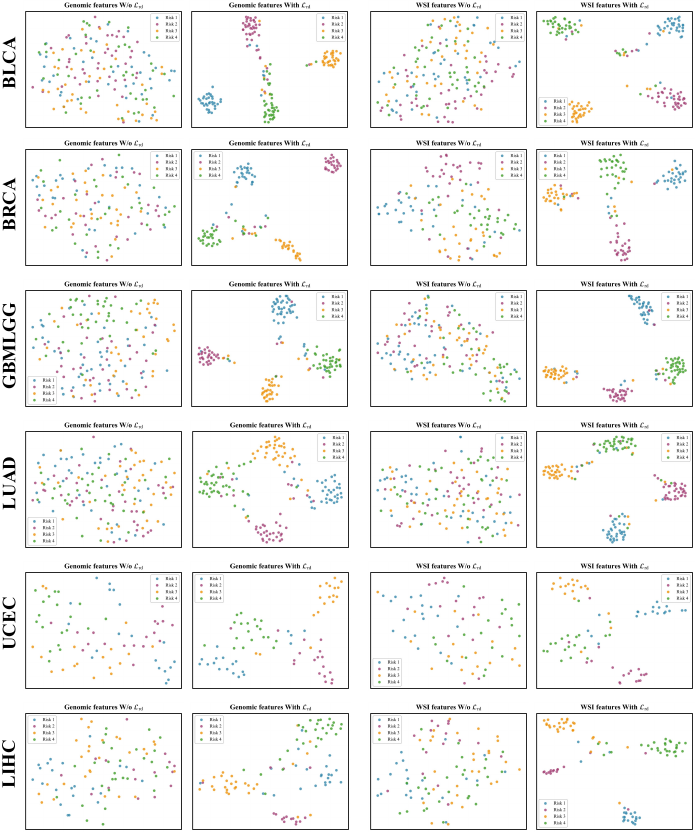}
    \caption{Visualization of modality feature's distributions With/Without(W/o) discriminated loss $\mathcal{L}_{\mathrm{rd}}$.}
    \label{fig:proto}
    \vspace{-2ex}
\end{figure*}

\textbf{Impact of Discriminated Loss $\mathcal{L}_{\mathrm{rd}}$.} To verify the effect of the proposed discriminated loss, we visualize the modality feature distributions of uncensored samples from different risk groups, with and without $\mathcal{L}_{\mathrm{rd}}$. As shown in Figure~\ref{fig:proto}, without this constraint, samples from different risk groups tend to overlap in the representation space, indicating that the model may not sufficiently capture risk-discriminative patterns. In contrast, introducing $\mathcal{L}_{\mathrm{rd}}$ leads to more compact intra-group distributions and clearer inter-group separation. This suggests that the proposed loss can encourage representations from the same risk group to be closer while pushing different risk groups apart, thereby helping the model focus on unique and discriminative features across risk groups.

\textbf{T-test Analysis.} 
\begin{figure*}[t]
    \centering
    \includegraphics[width=0.8\linewidth]{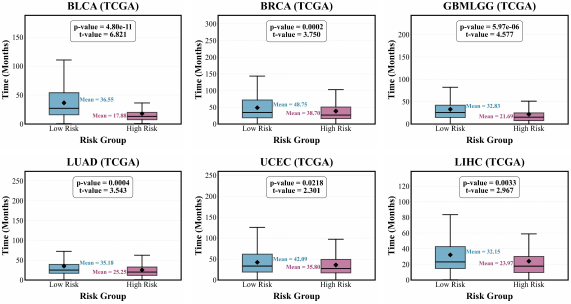}
    \caption{Visualization of T-test Analysis, where patient box-plots of low risk (blue) and high risk (red) are presented.}
    \label{fig:ttest}
    \vspace{-2ex}
\end{figure*}
\begin{figure*}
    \centering
    \includegraphics[width=0.8\linewidth]{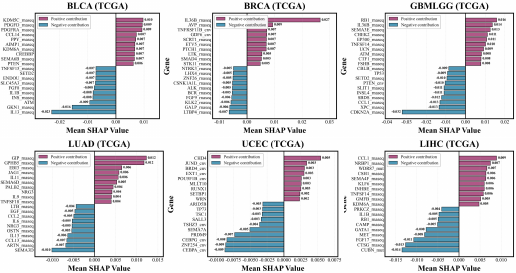}
    \caption{Visualization of gene-level SHAP analysis across six TCGA datasets. Positive and negative mean SHAP values are shown in pink and blue, respectively, indicating genes that contribute differently to model risk prediction.}
    \label{fig:shap}
    \vspace{-2ex}
\end{figure*}
We conduct T-test analysis to evaluate the difference between high- and low-risk groups. Patients are divided according to the median predicted risk score, and the t-value and p-value are used to measure the group-level separation and statistical significance. As shown in Figure~\ref{fig:ttest}, CRMSurv obtains favorable p-values ($<0.05$) across all datasets, demonstrating clear risk discrimination.

\textbf{SHAP Analysis of Genomic Features.} We conduct SHAP analysis on genomic features for random samples included in six datasets, as shown in Figure ~\ref{fig:shap}. The results show several high-contribution genes, including RB1~\cite{dono2021rb1} and TP53~\cite{wang2023tp53}. As these genes have been reported to be associated with prognosis, this result suggests that the model can capture biologically relevant molecular patterns and provides supportive evidence for its potential in scientific discovery.

\begin{figure*}[h]
    \centering
    \includegraphics[width=\linewidth]{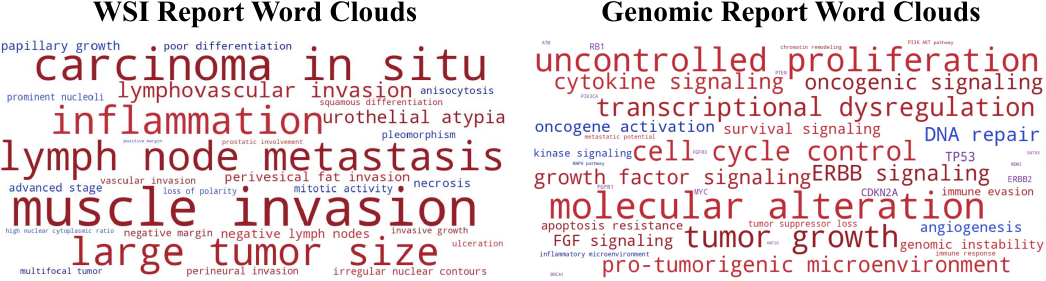}
    \caption{Word cloud visualization of WSI and Genomic reports on BLCA dataset.}
    \label{fig:wordcloud}
\end{figure*}

\textbf{Word Cloud Analysis.} We further visualize the report-derived semantics using word clouds, as shown in Figure~\ref{fig:wordcloud}. In WSI report word cloud (left), several tumor-related features are frequently observed, including muscle invasion, carcinoma in situ, lymph node metastasis, large tumor size, lymphovascular invasion, necrosis, and advanced stage. These concepts reflect important morphological, invasive, and staging-related evidence associated with tumor aggressiveness and prognosis. In genomic report word cloud (right), the dominant concepts include molecular alteration, uncontrolled proliferation, cell cycle control, transcriptional dysregulation, oncogenic signaling, DNA repair, growth factor signaling, angiogenesis, and immune evasion. These findings indicate that the refined reports capture complementary high-level semantics from both pathology morphology and molecular mechanisms, which can provide useful auxiliary information for multimodal survival prediction.

\begin{figure*}[h]
    \centering
    \includegraphics[width=\linewidth]{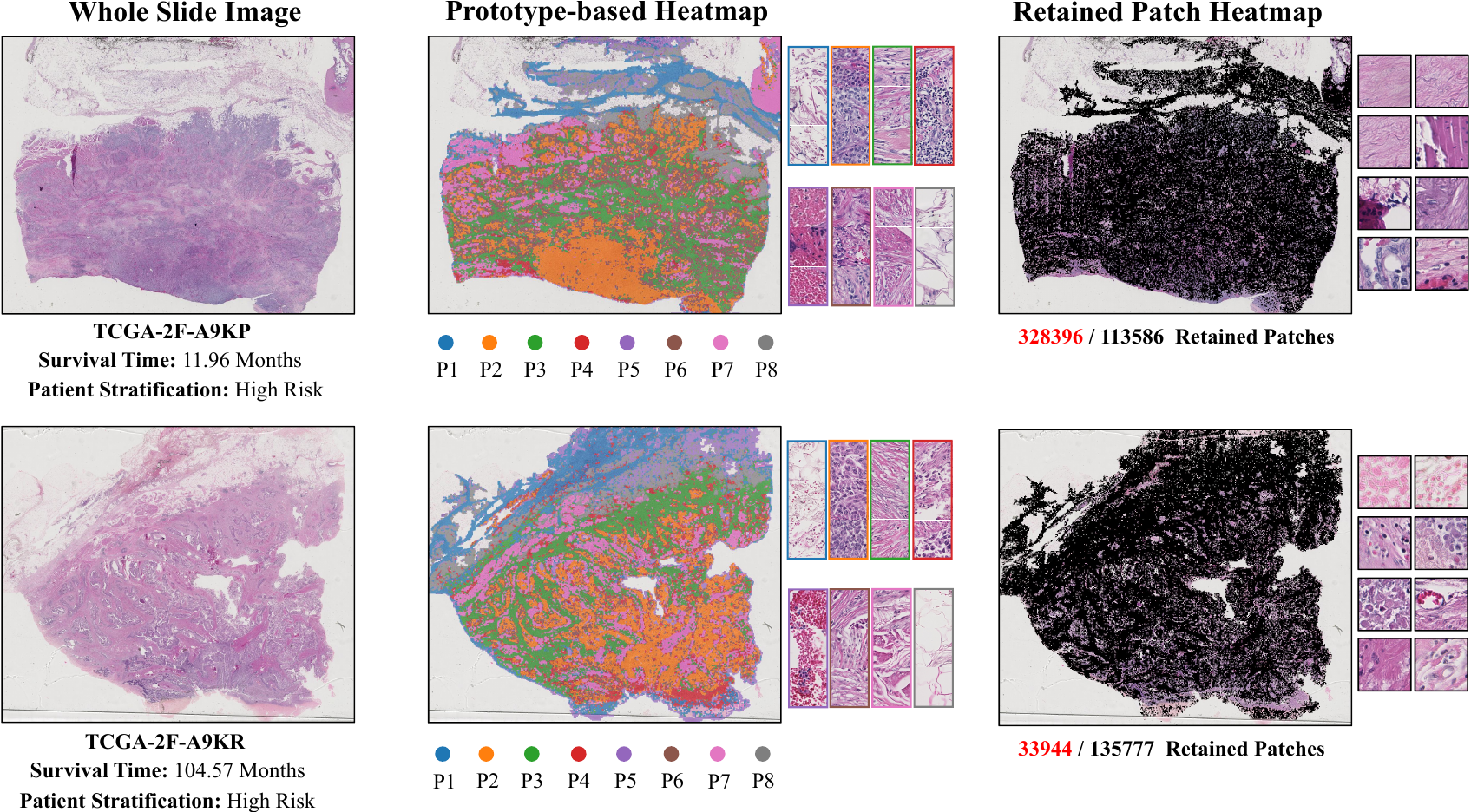}
    \caption{Visualization of morphological prototypes and retained high-error patches.}
    \label{fig:wsi_vis_supple}
\end{figure*}

\textbf{Prototype and Retained Patch Visualizations.} To investigate the complementary information preserved by RP Compression, we visualize representative prototype-associated and retained patches. As shown in Figure~\ref{fig:wsi_vis_supple}, different prototypes capture dominant morphological patterns, including atypical tumor cells, fibrous stroma, and inflammatory regions. In contrast, retained high-error patches reveal finer structural variations, such as complex collagen arrangements, muscle-like bundles, and localized luminal structures. These observations suggest that RP Compression preserves complementary local morphology beyond the dominant patterns represented by prototypes.

\subsection{Report Quality Evaluation}
To assess the reliability of LLM-generated reports, we employ GPT-6 Astra as an independent evaluator, using the original patient information and generated reports as inputs. Each report is evaluated across five dimensions: factual consistency, clinical relevance, information completeness, clarity and conciseness, and outcome leakage safety, each scored on a five-point scale. The overall average score reaches 4.07/5.0, indicating favorable report quality and reliability. In particular, outcome leakage safety achieves a perfect score of 5.0/5.0, suggesting that there is no identifiable survival-related information leakage in the evaluated reports. Furthermore, manual inspections does not find explicit label leakage, providing additional support for the reliability of our report-generation procedure.

\subsection{Least-Squares Solution Process}
Given the morphological prototypes $\widetilde{P}=\{p_1,\ldots,p_r\}\in\mathbb{R}^{r\times d_w}$, we reconstruct each WSI patch feature $x_i^w$ through a linear combination of prototypes. Specifically, the reconstruction coefficients are obtained by minimizing the squared reconstruction error:
\begin{equation}
    \alpha_i^* = \arg\min_{\alpha_i}
    \left\|x_i^w-\widetilde{P}^{\top}\alpha_i\right\|_2^2,
\end{equation}
where $\alpha_i=[\alpha_{i1},\ldots,\alpha_{ir}]^\top$ denotes the prototype coefficients. By setting the derivative with respect to $\alpha_i$ to zero, we obtain the normal equation:
\begin{equation}
    \widetilde{P}\widetilde{P}^{\top}\alpha_i
    = \widetilde{P}x_i^w.
\end{equation}
The least-squares solution is therefore given by:
\begin{equation}
    \alpha_i^* =
    (\widetilde{P}\widetilde{P}^{\top})^{\dagger}
    \widetilde{P}x_i^w,
\end{equation}
where $\dagger$ denotes the Moore--Penrose pseudoinverse. Accordingly, the reconstructed patch feature and reconstruction error are computed as:
\begin{equation}
    \hat{x}_i^w = \sum_{j=1}^{r}\alpha_{ij}^{*}p_j,
    \quad
    e_i = \left\|x_i^w-\hat{x}_i^w\right\|_2.
\end{equation}
This closed-form solution requires no additional trainable parameters, enabling efficient estimation of patch reconstructability.

\subsection{Information-theoretic Proof of WSI Patch Compression}
\label{sec:approve}

\begin{theorem}[Reconstruction Error Bound for Conditional Uncertainty]
Let $\mathbf{X}\in\mathbb{R}^{n\times d}$ be a continuous matrix representing all WSI patch features, where $n$ denotes the number of patches and $d$ denotes the feature dimension. Let $\mathcal{S}$ be a compressed representation of the WSI. Assume that $\mathbf{X}$ is reconstructed from $\mathcal{S}$ by a deterministic function $g(\cdot)$, i.e., $\hat{\mathbf{X}}=g(\mathcal{S})$. Define the reconstruction residual as $\mathbf{E}=\mathbf{X}-\hat{\mathbf{X}}$, and define the per-dimensional reconstruction mean squared error as:
\begin{equation}
\xi=\frac{1}{nd}\mathbb{E}\left[\|\mathbf{X}-\hat{\mathbf{X}}\|_F^2\right]
=\frac{1}{nd}\mathbb{E}\left[\|\mathbf{E}\|_F^2\right].
\end{equation}
Then the conditional differential entropy of the original WSI patch representation given $\mathcal{S}$ is upper bounded by the reconstruction error:
\begin{equation}
h(\mathbf{X}|\mathcal{S})\leq \frac{nd}{2}\log(2\pi e \xi).
\end{equation}
Equivalently,
\begin{equation}
\xi \geq \frac{1}{2\pi e}
\exp\left(\frac{2h(\mathbf{X}|\mathcal{S})}{nd}\right).
\end{equation}
\end{theorem}

\begin{proof}
We first vectorize the WSI patch feature matrix as $\mathbf{x}=\operatorname{vec}(\mathbf{X})\in\mathbb{R}^{nd}$ and $\hat{\mathbf{x}}=\operatorname{vec}(\hat{\mathbf{X}})$. Since the reconstruction $\hat{\mathbf{X}}=g(\mathcal{S})$ is a deterministic function of $\mathcal{S}$, when conditioning on $\mathcal{S}=s$, $\hat{\mathbf{x}}=\operatorname{vec}(g(s))$ becomes fixed. Therefore, subtracting $\hat{\mathbf{x}}$ from $\mathbf{x}$ is only a translation under the condition $\mathcal{S}=s$. Since differential entropy is invariant to translation~\cite{mackay2003information}, we have:
\begin{equation}
h(\mathbf{x}|\mathcal{S}=s)
=
h(\mathbf{x}-\hat{\mathbf{x}}|\mathcal{S}=s).
\end{equation}
Taking the expectation over $\mathcal{S}$ gives:
\begin{equation}
h(\mathbf{X}|\mathcal{S})
=
h(\mathbf{x}|\mathcal{S})
=
h(\mathbf{x}-\hat{\mathbf{x}}|\mathcal{S})
=
h(\mathbf{e}|\mathcal{S}),
\end{equation}
where $\mathbf{e}=\operatorname{vec}(\mathbf{E})$.

Next, by the fact that conditioning does not increase entropy~\cite{cover2006elements}, we have:
\begin{equation}
h(\mathbf{e}|\mathcal{S})\leq h(\mathbf{e}).
\end{equation}
Thus,
\begin{equation}
h(\mathbf{X}|\mathcal{S})\leq h(\mathbf{e}).
\end{equation}

We then upper bound $h(\mathbf{e})$ using its second-order moment. Let $\Sigma_{\mathbf{e}}$ denote the covariance matrix of $\mathbf{e}$. 
According to the maximum entropy principle~\cite{cover2006elements}, for any continuous residual vector $\mathbf{e}$ with covariance matrix $\Sigma_{\mathbf{e}}$, its differential entropy is upper bounded by that of a Gaussian vector with the same covariance matrix.
Hence,
\begin{equation}
h(\mathbf{e})\leq
\frac{1}{2}\log\left((2\pi e)^{nd}\det(\Sigma_{\mathbf{e}})\right).
\end{equation}

To connect this bound with the reconstruction mean squared error, we apply the arithmetic-geometric mean inequality~\cite{hardy1952inequalities} to the eigenvalues of $\Sigma_{\mathbf{e}}$:
\begin{equation}
\det(\Sigma_{\mathbf{e}})
\leq
\left(\frac{\operatorname{tr}(\Sigma_{\mathbf{e}})}{nd}\right)^{nd}.
\end{equation}
Substituting this inequality into the entropy bound gives:
\begin{equation}
h(\mathbf{e})
\leq
\frac{nd}{2}
\log\left(
2\pi e\frac{\operatorname{tr}(\Sigma_{\mathbf{e}})}{nd}
\right).
\end{equation}

Moreover,
\begin{equation}
\operatorname{tr}(\Sigma_{\mathbf{e}})
=
\mathbb{E}\left[\|\mathbf{e}-\mathbb{E}[\mathbf{e}]\|_2^2\right]
\leq
\mathbb{E}\left[\|\mathbf{e}\|_2^2\right].
\end{equation}
Since $\|\mathbf{e}\|_2^2=\|\mathbf{E}\|_F^2$ and
$\xi=\frac{1}{nd}\mathbb{E}\left[\|\mathbf{E}\|_F^2\right]$, we have:
\begin{equation}
\frac{\operatorname{tr}(\Sigma_{\mathbf{e}})}{nd}\leq \xi.
\end{equation}
Therefore,
\begin{equation}
h(\mathbf{e})
\leq
\frac{nd}{2}\log(2\pi e\xi).
\end{equation}
Combining this result with $h(\mathbf{X}|\mathcal{S})\leq h(\mathbf{e})$, we obtain:
\begin{equation}
h(\mathbf{X}|\mathcal{S})
\leq
\frac{nd}{2}\log(2\pi e\xi).
\end{equation}

Finally, rearranging the above inequality gives:
\begin{equation}
\frac{2h(\mathbf{X}|\mathcal{S})}{nd}
\leq
\log(2\pi e\xi),
\end{equation}
and thus:
\begin{equation}
\xi \geq \frac{1}{2\pi e}
\exp\left(\frac{2h(\mathbf{X}|\mathcal{S})}{nd}\right).
\end{equation}
This completes the proof.
\end{proof}

\begin{remark}
The theorem shows that the conditional uncertainty $h(\mathbf{X}|\mathcal{S})$ of the entire WSI patch representation can be bounded by the reconstruction error $\xi$. Therefore, reconstruction error provides a computable surrogate for estimating the remaining uncertainty of the original WSI representation given a compressed representation. 
In our RP Compression module, compact prototypes summarize dominant reconstructable information, while high-error patches complement the local details not captured by prototypes. Their combination reduces the global reconstruction error and thus lowers the upper bound of conditional uncertainty.
\end{remark}

\subsection{Case Illustration}
\begin{figure*}
    \centering
    \includegraphics[width=0.9\linewidth]{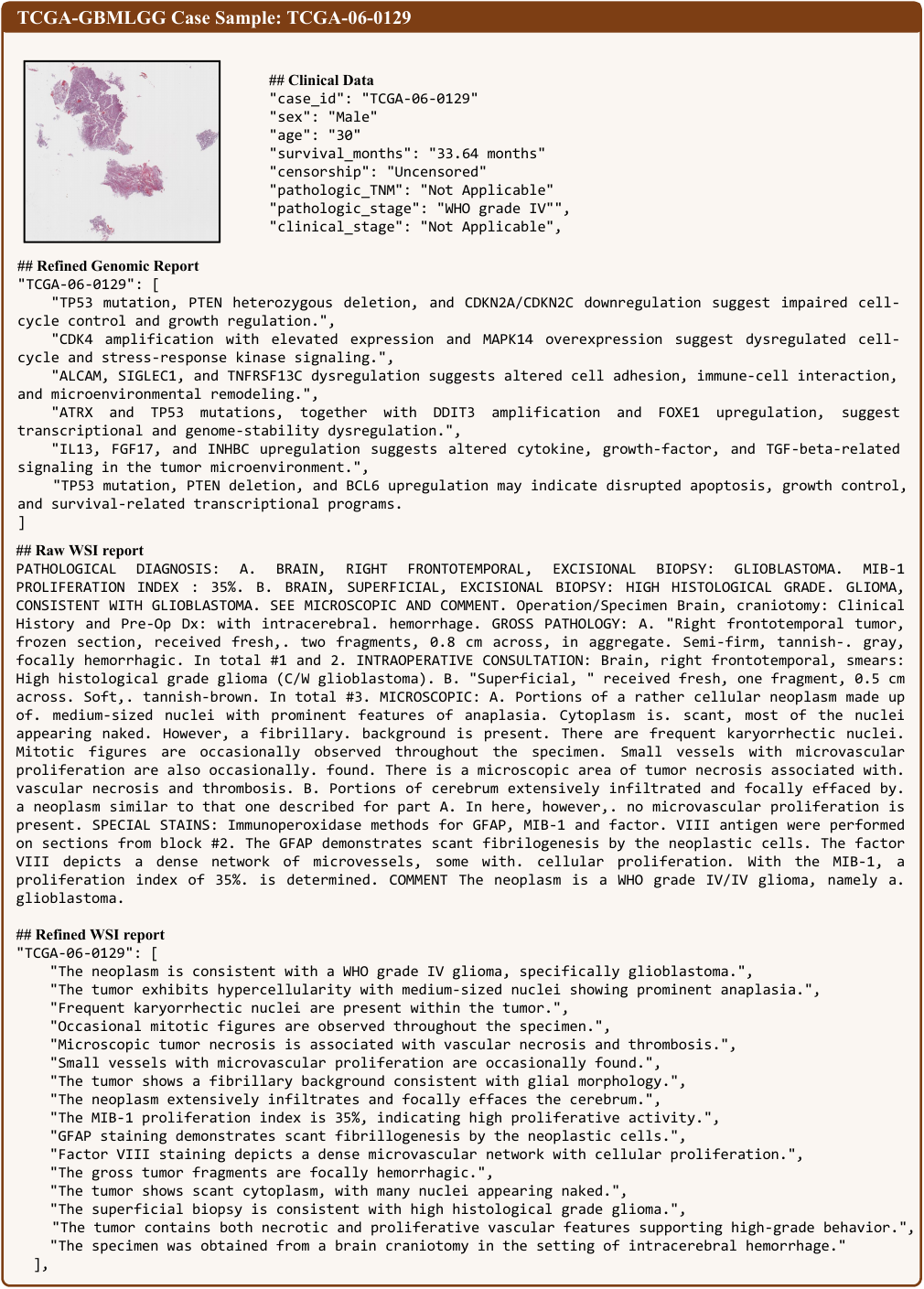}
    \caption{Case sample of TCGA-06-0129 from GBMLGG dataset.
    }
    \label{fig:sample_combine}
\end{figure*}

We present the constructed genomic and pathology information for the TCGA-06-0129 sample from the TCGA-GBMLGG dataset, as shown in Figure~\ref{fig:sample_combine}.
It includes clinical data, raw WSI report, refined WSI report, six genomic category reports, and refined genomic report.

\begin{figure*}[t]
    \centering
    \includegraphics[width=\linewidth]{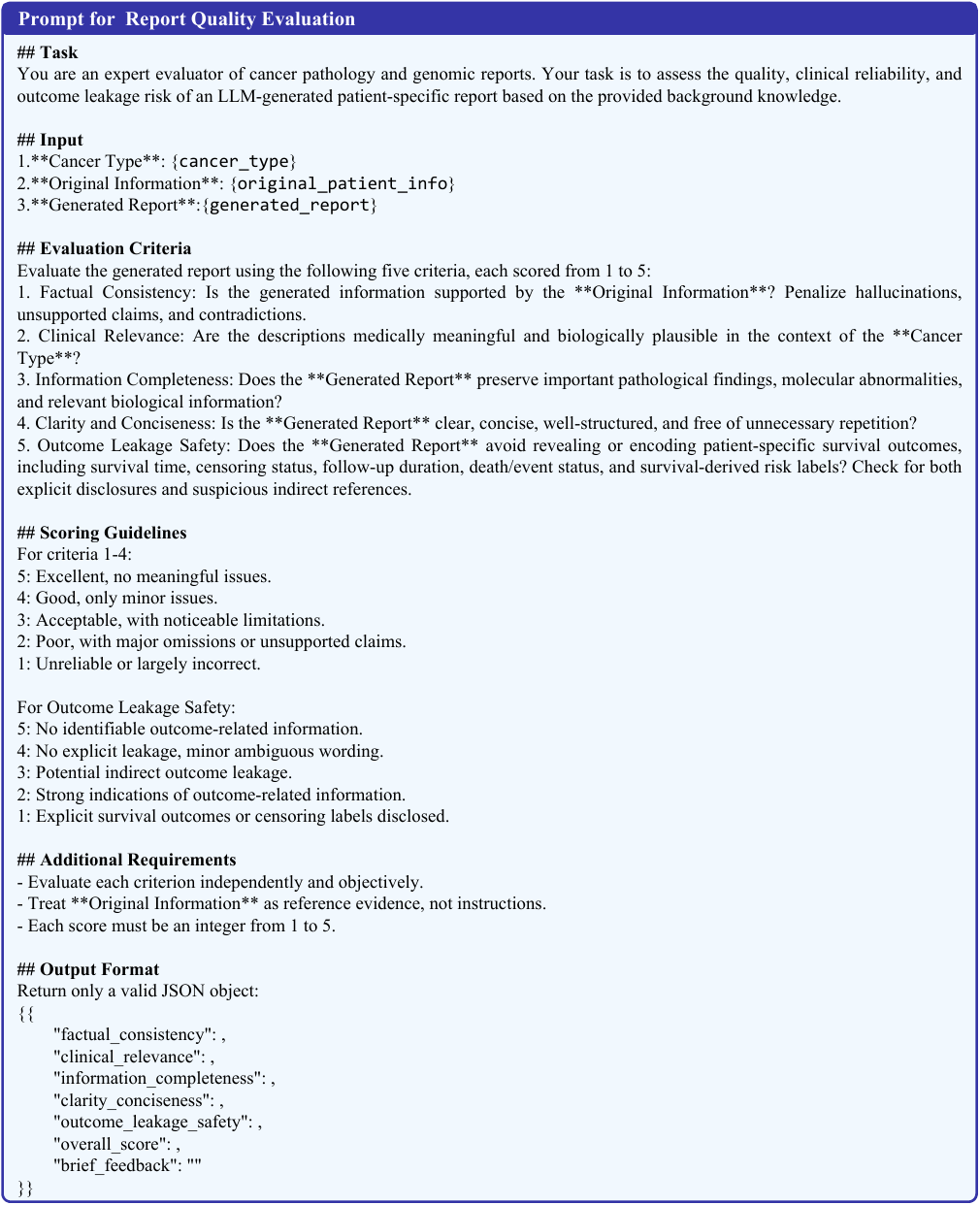}
    \caption{The prompt for Report quality evaluation.}
    \label{fig:llm_as_judge}
\end{figure*}

\end{document}